%% file: acl_latex.tex
\pdfoutput=1

\documentclass[11pt]{article}

\usepackage{longtable}
\usepackage{graphicx}
\usepackage{subcaption}
\usepackage{amsmath}
\usepackage{stackengine}

\usepackage{tabularx}
\usepackage{multirow}
\usepackage{booktabs}
\usepackage{url}
\usepackage{xcolor}
\usepackage{authblk}

\usepackage{ACL2023}
\usepackage{times}
\usepackage{latexsym}

\usepackage[T1]{fontenc}

\usepackage[utf8]{inputenc}

\usepackage{microtype}

\usepackage{inconsolata}

\newcommand{\dataset}{\textsc{DMaste }} 
\newcommand{\datasetwb}{\textsc{DMaste}}

\title{Measuring Your ASTE Models in The Wild:  A Diversified Multi-domain Dataset For Aspect Sentiment Triplet Extraction}

\author{Ting Xu\textsuperscript{\rm $\spadesuit$}\;,
 Huiyun Yang\textsuperscript{\rm $\clubsuit$}, 
 Zhen Wu\textsuperscript{\rm $\spadesuit$}\;, 
 Jiaze Chen\textsuperscript{\rm $\clubsuit$}, 
 Fei Zhao\textsuperscript{\rm $\spadesuit$},
 Xinyu Dai\textsuperscript{\rm $\spadesuit$}\\
\textsuperscript{\rm $\spadesuit$}National Key Laboratory for Novel Software Technology, Nanjing University\\
\textsuperscript{\rm $\clubsuit$}ByteDance\\
\texttt{\{xut, zhaof\}@smail.nju.edu.cn}, \texttt{\{wuz, daixinyu\}@nju.edu.cn}\\
\texttt{\{yanghuiyun.11, chenjiaze\}@bytedance.com}
}

\begin{document}
\maketitle
\input{chapters_0515/main.tex}

\end{document}

%% file: chapters_0515/main.tex
\input{chapters_0515/0_abstract}
\input{chapters_0515/1_intro}

\input{chapters_0115/2_related_work}
\input{chapters_0515/3_dataset}

\input{chapters_0115/4_experiment_settings/main.tex}

\input{chapters_0115/5_results/main.tex}

\input{chapters_0115/6_conclusion}
\input{chapters_0115/7_limitations.tex}
\input{chapters_0515/8_acknowledgement}
\bibliography{anthology,custom}
\bibliographystyle{acl_natbib}

\appendix

\input{chapters_0115/8_appendix/main.tex}

%% file: chapters_0515/0_abstract.tex
\begin{abstract}
    Aspect  Sentiment Triplet Extraction (ASTE) is widely used in various applications. However, existing ASTE datasets are limited in their ability to represent real-world scenarios, hindering the advancement of research in this area. In this paper, we introduce a new dataset, named \datasetwb, which is manually annotated to better fit real-world scenarios by providing more diverse and realistic reviews for the task. The dataset includes various lengths, diverse expressions, more aspect types, and more domains than existing datasets. We conduct extensive experiments on \dataset in multiple settings to evaluate previous ASTE approaches. Empirical results demonstrate that \dataset is a more challenging ASTE dataset. Further analyses of in-domain and cross-domain settings provide promising directions for future research. Our code and dataset are available at \href{https://github.com/NJUNLP/DMASTE}{https://github.com/NJUNLP/DMASTE}.

\end{abstract}

%% file: chapters_0515/1_intro.tex
\section{Introduction}
\label{sec:intro}

\begin{figure}[htbp] 
\centering
\includegraphics[width=0.45\textwidth]{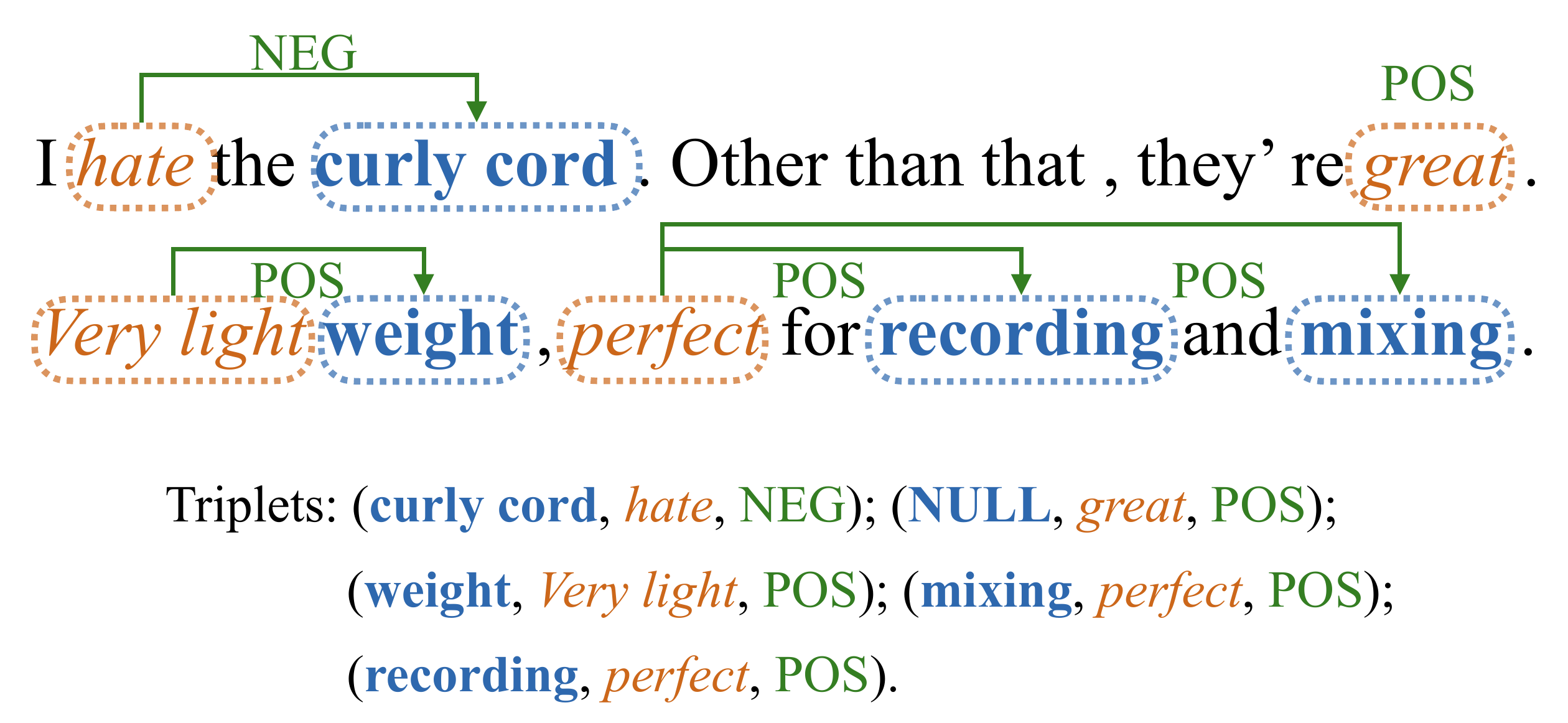} 
\caption{An example of the ASTE task. The terms highlighted in blue are aspect terms. The terms in orange are opinion terms. The words in green are sentiment polarities. "NULL" denotes the implicit aspect.} 
\label{fig:intro-example} 
\end{figure}

Aspect sentiment triplet extraction \citep[ASTE;][]{peng2020knowing}, a fine-grained task in sentiment analysis \citep{hussein2018survey}, has attracted considerable interest recently \citep{peng2020knowing, xu-etal-2020-position}.  The objective of this task is to extract the sentiment triplet, comprising of an aspect term, an opinion term, and a sentiment polarity, from a given review. As depicted in Figure \ref{fig:intro-example}, an example of the sentiment triplet is ({\textit{"curly cord"}, \textit{"hate"}, \textit{NEG}}), representing a \textit{negative} sentiment toward the aspect term \textit{"curly cord"} using the opinion term {\textit{"hate"}}. The ASTE task requires a deep understanding of linguistic forms and structures (e.g., aspect terms are usually nouns or verbs used as subjects or objects in a sentence), as well as the ability to identify the relationships between the various linguistic components (e.g.,  how to pair the aspect terms and opinion terms) in a given text.

Prior ASTE methods \citep{yan-etal-2021-unified, xu-etal-2021-learning} have achieved promising results on existing academic datasets \citep{peng2020knowing, xu-etal-2020-position, wu-etal-2020-grid}, greatly promoting the development and application of ASTE. 
However, the datasets employed in these studies remain comparatively uncomplicated, leading to disparities between these datasets and real-world settings in terms of various factors, such as length, expression diversity, domain distribution, etc.
For instance, most reviews in existing datasets are of short length, with an average of 16 words per review, while reviews in real-world scenarios are longer (an average of more than 50 words). Additionally,  expressions used in these datasets are typically simple and straightforward, with limited diversity in lexicality and syntactic. Furthermore, existing datasets typically contain two domains, i.e., restaurant and laptop, with very limited domain distributions. In a nutshell, these gaps hide the complexity of real-world scenarios, and therefore, impede the exploration to fully understand and address the challenges presented in real-world ASTE tasks.



\input{figures/fig-data-cmp.tex}

\input{tables/data_comparison_new.tex}

In order to bridge the gap and better simulate real-world scenarios, we  create a new dataset, named Diversified Multi-domain ASTE (\datasetwb), which is manually annotated to provide a more diverse and realistic set of reviews for the task. As Table~\ref{tab:cmp} and Figure~\ref{fig:intro-cmp} shows, the key characteristics of \dataset can be summarized as follows: (1) \textbf{Various lengths}: \dataset covers reviews of various lengths, ranging from 1 to 250 words, with an average of 59 words per review. (2) \textbf{Diverse expressions}: more part-of-speech and dependency n-grams show a wide variety of lexical bundles and syntactic in \datasetwb, 
which can better represent the complexity and diversity of real-world scenarios. (3) \textbf{More aspect types}: \dataset  includes triplets annotated with both implicit and explicit aspect terms, providing  a more comprehensive understanding of the target being discussed. (4) \textbf{More domains}: \dataset covers eight domains, enabling more comprehensive research in ASTE like single-source and multi-source domain adaptation.
To summarize, these characteristics make \dataset a better benchmark to verify the ability of ASTE approaches in real-world scenarios.
 
To thoroughly investigate the challenges introduced by \dataset and explore promising directions for future ASTE research, we implement several representative methods and empirically evaluate \dataset under multiple settings:
\begin{itemize}
    \item In-domain results show that the performance of current models declines significantly on \datasetwb. And analysis reveals that \textit{long reviews, complex sentences, and implicit aspect terms make \dataset a challenging dataset}.
    \item In the single-source domain adaptation setting, we 
    observe a positive correlation between transfer performance and domain similarity. But simply learning domain-invariant features may lead to the loss of task-specific knowledge, which suggests that \textit{reducing domain discrepancy while keeping the task-specific knowledge can be a future direction}.
    \item We observe the \textit{negative transfer} \citep{Rosenstein2005} in the multi-source domain adaptation setting, 
    and find the negative transfer occurs mainly in dissimilar source-target pairs. This indicates that \textit{the domain similarity may be a useful guideline for domain selection} in future multi-source cross-domain ASTE research.
\end{itemize}
  Overall, the results of \dataset on multiple settings provide a deeper understanding of the challenges and future directions for ASTE research. We believe this work will bring a valuable research resource and benchmark for the community.


%% file: figures/fig-data-cmp.tex
\begin{figure*}
     \centering
     \begin{subfigure}[b]{0.48\textwidth}
         \centering
         \includegraphics[trim=4 4 4 4,clip,width=\linewidth]{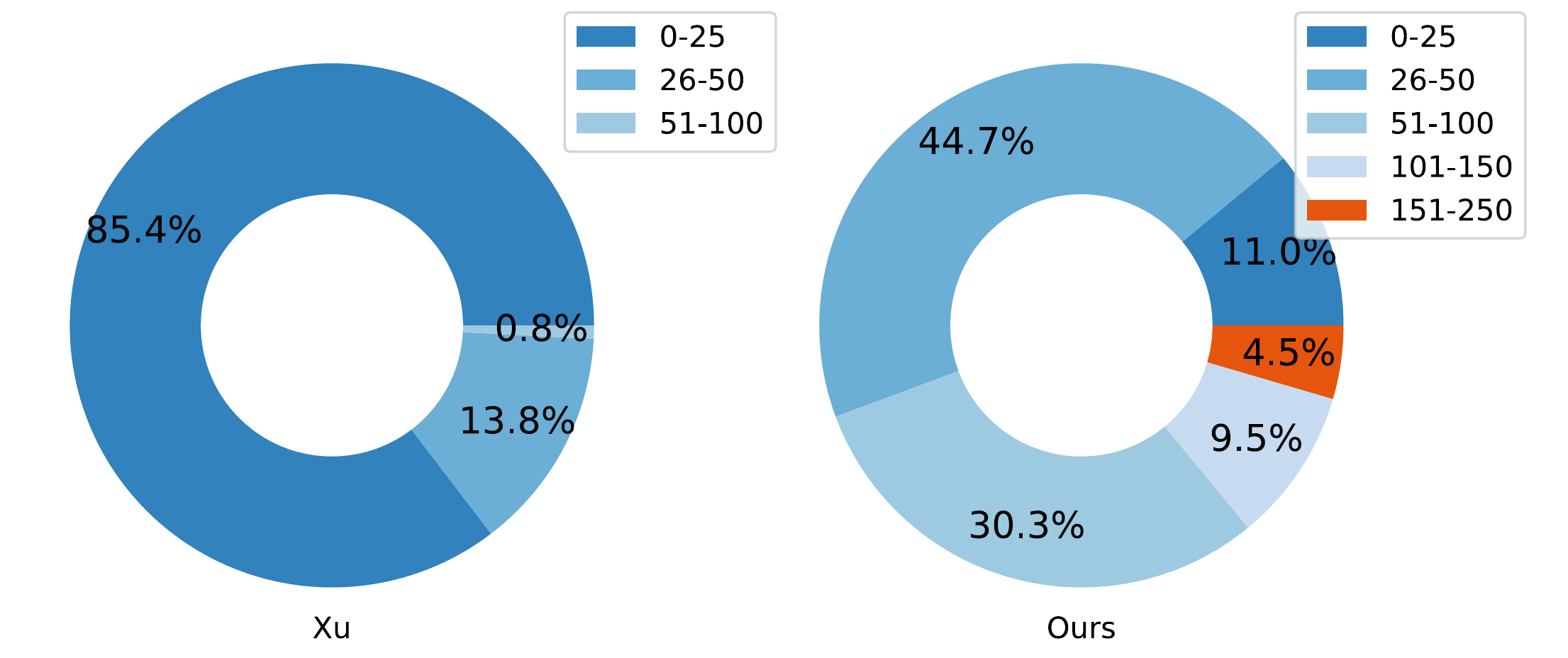}
         \caption{Distribution of review length.}
         \label{fig:intro-length}
     \end{subfigure}
     \begin{subfigure}[b]{0.48\textwidth}
         \centering
         \includegraphics[trim=4 4 4 4,clip,width=\linewidth]{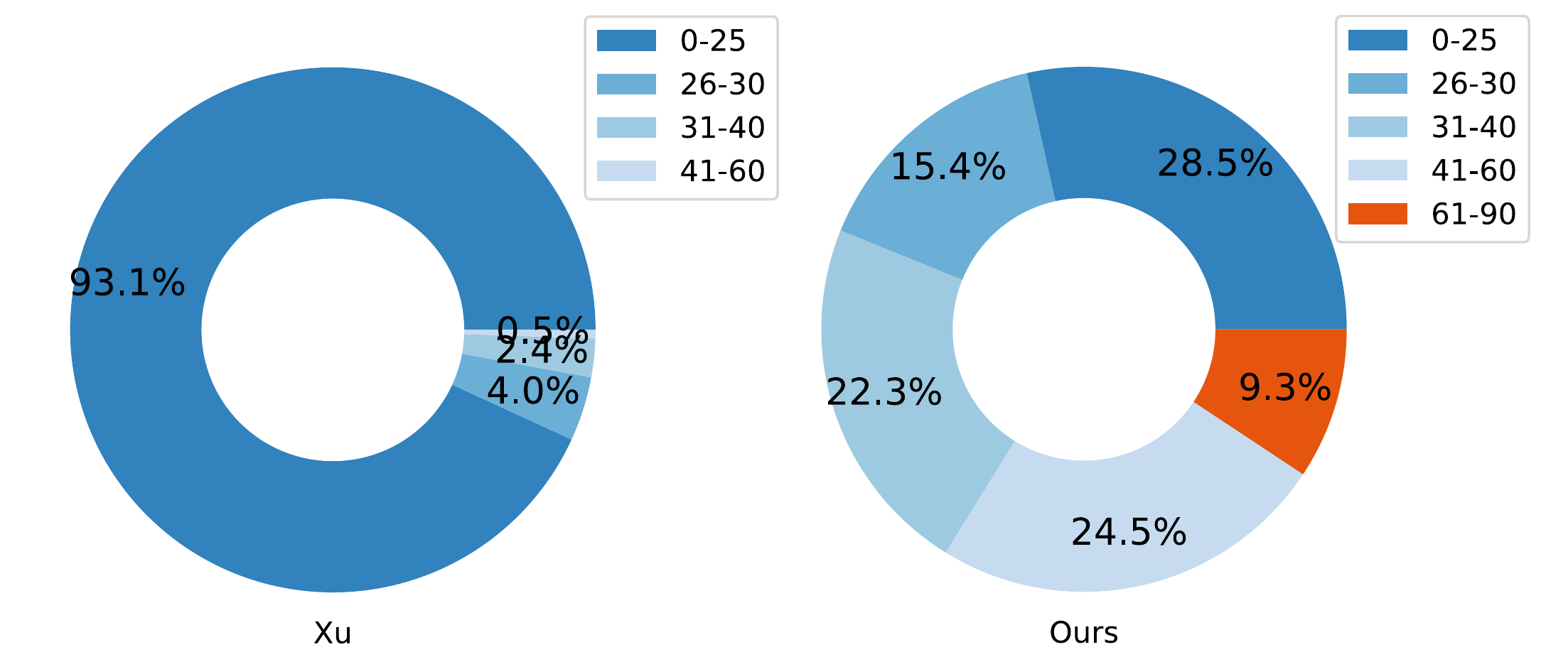}
         \caption{Distribution of the number of POS 2-gram.}
         \label{fig:intro-pos}
     \end{subfigure}
        \caption{Comparisons between \citet{xu-etal-2020-position} and our dataset on review length and the number of POS 2-grams per review. 
        }
        \label{fig:intro-cmp}
\end{figure*}

%% file: tables/data_comparison_new.tex
\begin{table*}
    \small
    \centering
    \newcolumntype{Z}{>{\centering\let\newline\\\arraybackslash\hspace{0pt}}X}
    \begin{tabular}{l c c c c c c c c c c c c c}
    
    \toprule
    Dataset & \#D & \#R & \#W/\#R & \#T & \#IA & \#T/\#I & POS \#n-gram & DP \#n-gram & \#vocab\\
    \midrule
    $\mathcal{D}_P$ & 2 & 6037 & 16.49 & 9309 & 0 & 1.54 & 14.25/16.05/16.52 & 13.51/14.24/14.43 & 6512\\
    $\mathcal{D}_W$ & 2 & 6009 & 16.46 & 10390 & 0 & 1.73 & 14.22/16.02/16.48 & 13.48/14.21/14.40 & 6484\\
    $\mathcal{D}_X$ & 2 & 5989 & 16.43 & 10252 & 0 & 1.71  & 14.20/16.00/16.46 & 13.46/14.19/14.38 & 6467\\
    \datasetwb & 8 & 7524 & 59.68 & 28233 & 11945 & 3.75 &36.67/52.11/57.92 &  35.08/40.20/41.93 &19226\\
    \bottomrule
    \end{tabular}
    \caption{Comparisons between \dataset and existing representative ASTE datasets, where $\mathcal{D}_P$, $\mathcal{D}_W$, $\mathcal{D}_X$ denote the datasets of \citet{peng2020knowing}, \citet{wu-etal-2020-grid}, and \citet{xu-etal-2020-position}.
    \#D, \#R, \#W, \#T, and \#IA denote the numbers of domains, reviews, words, triplets, and triplets with implicit aspect terms, respectively.
    \#n-gram denotes the numbers of 2/3/4-grams in part-of-speech sequences (POS) or dependency parsing trees (DP) per review. \#vocab denotes the size of the vocabulary for each dataset. }
    \label{tab:cmp}
\end{table*}

%% file: chapters_0115/2_related_work.tex
\section{Related Work}
\subsection{ASTE Datasets}
Current ASTE datasets \citep{peng2020knowing, xu-etal-2020-position, wu-etal-2020-grid} share a common origin and are constructed through similar processes. Specifically, they originate from SemEval Challenge datasets \citep{pontiki-etal-2014-semeval, pontiki2015semeval, pontiki2016semeval}, which provide aspect terms and corresponding sentiments for reviews in the domains of restaurant and laptop.  Based on the datasets,  \citet{fan2019target} annotate the aspect-opinion pairs. 
To provide more detailed information about the review text, researchers resort to extracting sentiment triplets from the review, i.e., aspect term, opinion term, and sentiment polarity. \citet{peng2020knowing}  and \citet{wu-etal-2020-grid}  construct the ASTE datasets by aligning the aspect terms between the datasets of \citet{fan2019target} and original SemEval Challenge datasets. As noted by \citet{xu-etal-2020-position}, the dataset of \citet{peng2020knowing} does not contain cases where one opinion term is associated with multiple aspect terms. \citet{xu-etal-2020-position} subsequently refine the dataset and release a new version. 

However, all of these datasets contains reviews of limited diversity from only two domains. Additionally, they all require aspect terms to align the aspect-sentiment pair and aspect-opinion pair, thus they do not include implicit aspect terms \citep{poria2014rule}. Our dataset, \datasetwb, addresses these limitations by providing a more diverse set of reviews covering more domains and annotate triplets with both implicit and explicit aspect terms, making it better suited for real-world scenarios \footnote{\citet{cai-etal-2021-aspect, zhang-etal-2021-aspect-sentiment} can also be transformed into sentiment triplets. We omit the comparison with them since they share almost the same data origination with the existing triplet dataset.}.

\subsection{ASTE Methods}
Corresponding solutions for ASTE can be divided into three categories: tagging-based \citep{li2019unified, peng2020knowing, xu-etal-2020-position, zhang-etal-2020-multi-task, wu-etal-2020-grid, xu-etal-2021-learning}, MRC-based \citep{chen2021bidirectional, mao2021joint} and generation-based \citep{zhang2021towards, yan-etal-2021-unified, fei2021nonautoregressive, mukherjee-etal-2021-paste}. 
The tagging-based method employs a sequence or grid tagging framework to extract the aspect and opinion terms, then combines them to predict the sentiment. 
The MRC-based method constructs a specific query for each factor in the triplet and extracts them through the answer to the query.
The generation-based method transforms the ASTE task into a sequence generation problem and employs sequence-to-sequence (seq2seq) models. Then it decodes the triplets through a specifically designed algorithm. In this paper, we employ some representative methods in three categories and explore \dataset in multiple settings.


%% file: chapters_0515/3_dataset.tex
\input{tables/data_statistics.tex}

\section{Dataset}
\label{sec:dataset}

To construct a dataset that is more representative of real-world scenarios, we manually annotate a new dataset, named Diversified Multi-domain ASTE (\datasetwb).  In this section, we first present a detailed description of the data collection and annotation process. Then we demonstrate the superiority of \datasetwb, through a comparison with previous datasets in terms of key statistics and characteristics.

\subsection{Collection}
The data collection process of \dataset is carried out in three stages:
(1) We select the Amazon dataset \citep{ni-etal-2019-justifying} as our source of data due to its large volume of reviews from various regions around the world, which aligns with the goal of creating a dataset that is more representative of real-world scenarios.
(2) We select four of the most popular domains from the Amazon dataset (Appendix \ref{sec:app-dataset}), and randomly sample a portion of the data for annotation.
(3) To further enable comprehensive exploration of ASTE like domain adaptation settings, we additionally sample four additional domains and annotate a smaller portion of the data for testing in domain adaptation settings.

\subsection{Annotation}
\label{sec:annotation}

\paragraph{Simplified Annotation Guidelines \footnote{Due to the space limitation, we present detailed labeling guidelines in Appendix \ref{sec:app-guideline}.}.}Following \citet{peng2020knowing,xu-etal-2020-position,wu-etal-2020-grid}, we annotate \textit{(aspect term, opinion term, sentiment polarity)} triplets for each review, which may include multiple sentences.
An \textit{aspect term} refers to a part or an attribute of products or services.
Sometimes, the aspect term may not appear explicitly in the instance, i.e., implicit aspect terms \citep{poria2014rule}. We keep the triplets with implicit aspect terms.
An \textit{opinion term} is a word or phrase that expresses opinions or attitudes toward the aspect term.
A \textit{Sentiment polarity} is the sentiment type of the opinion term, which is divided into three categories: positive, negative, and neutral.
It is worth noting that one aspect term can be associated with multiple opinion terms and vice versa. 

\paragraph{Annotation Process.} 
In order to ensure the quality of the annotation, we employed 14 workers (annotators) to perform the annotation and 4 workers (verifiers) to perform 
 the quality-assurance sampling. Both groups are compensated based on the number of annotations. The annotation process is carried out using a train-trial-annotate-check procedure.
(1) \textit{Train}: All workers are trained on the task of annotation.
(2) \textit{Trial}: All workers try to annotate a small portion of the data to familiarize themselves with the task and to receive feedback on their annotations. Workers are required to reach 95\% accuracy in labeling before proceeding to the next phase. Following previous work in data annotation for ABSA \citep{barnes-etal-2018-multibooked}, we measure the inter-annotator agreement (IAA) using the AvgAgr metric \citep{esuli-etal-2008-annotating}. The IAA scores of annotated aspect terms, opinion terms, and sentiment triplets at the trial stage are shown in Table \ref{tab:iaa}. These scores are slightly lower than previous ABSA datasets \citep{barnes-etal-2018-multibooked}, which can be attributed to the inherent complexity of \datasetwb. (3) \textit{Annotate}: The annotators are responsible for annotating the data on a daily basis.
(4) \textit{Check}: The verifiers sampled 20\% of the annotations each day to ensure accuracy. If the accuracy of an annotator is found to be below 95\%, the data annotated by that worker for that day would be re-done until meeting the accuracy requirement. Otherwise, the annotations are accepted. To avoid false positives by the verifiers, we introduce an appeals mechanism. Please refer to the appendix \ref{sec:app-guideline} for detailed information.
\begin{table}[]
    \centering
    \small
    \begin{tabular}{cccc}
    \toprule
    Term    & Aspect terms & Opinion terms & Triplets\\
    \midrule
    IAA  & 0.666    & 0.664 & 0.593\\
    \bottomrule
    \end{tabular}
    \caption{Inter-annotator agreement (IAA) of aspect terms, opinion terms, and sentiment triplets.}
    \label{tab:iaa}
\end{table}

\subsection{Statistics and Characteristics}

To provide a deeper understanding of our proposed dataset, we present a series of statistics and characteristics in this section.

\paragraph{Various Lengths.} Figure \ref{fig:intro-length} illustrates the comparison of review length between \citet{xu-etal-2020-position} and \datasetwb. We find that \dataset contains reviews of more varied lengths. And Table \ref{tab:cmp} demonstrates that the average length of \dataset is 3.6 times that of \citet{xu-etal-2020-position}. 

\paragraph{Diverse Expressions.} We quantify the expression diversity by counting the vocabulary size, n-grams in part-of-speech (POS) sequences and dependency parsing (DP) trees \footnote{For the DP tree, we treat the parent node and child node as neighbors.}.  Figure \ref{fig:intro-pos} shows the comparison of POS 2-grams between \citet{xu-etal-2020-position} and \datasetwb. The results indicate that \dataset contains reviews with diverse expressions.  And Table \ref{tab:cmp} shows that the number of n-grams of POS and DP trees in \dataset is significantly higher than those of other datasets, with the number of 4-grams being about 3 times that of other datasets. Besides, the vocabulary size of \dataset is about 3 times that of other datasets. These statistics indicate that the reviews in \dataset are more diverse and complex.

\paragraph{More Aspect Types.} Table \ref{tab:cmp} shows that \dataset not only contains more explicit aspect terms than existing datasets but also includes annotations of implicit aspect terms, which constitute a large proportion. This provides a more comprehensive understanding of the target being discussed.
\paragraph{More Domains.} Table \ref{tab:cmp} illustrates that our dataset comprises four times the number of domains compared to existing datasets. Furthermore, the first four domains presented in Table \ref{tab:statistics}, which are characterized by a larger amount of annotated data, represent leading fields in e-commerce platforms (Appdenix \ref{sec:app-dataset}), providing a more realistic representation of popular topics. Additionally, we include an additional four domains with less annotated data to enable a more comprehensive analysis of the domain adaptation setting.

In summary, \dataset is a more realistic and diversified dataset, providing a suitable testbed to verify the ability of ASTE methods in real-world scenarios.


%% file: tables/data_statistics.tex
     

\begin{table*}[t]
    \small
    \centering
    \newcolumntype{Z}{>{\centering\let\newline\\\arraybackslash\hspace{0pt}}X}
     \begin{tabular}{l c c c c c c c c}
     
    \toprule
     Domain           &Electronics     & Fashion   & Beauty    &Home    &Book   &Pet    &Toy    &Grocery\\
    \midrule 
    Train   &   1395 & 851  & 535   & 1050  & 0     & 0     & 0     & 0 \\
    Dev     &   200  & 121  & 77    & 152   & 159   & 167   & 173   & 173\\
    Test    &   399  & 245  & 154   & 301   & 325   & 340   & 354   & 353\\
    \bottomrule
    \end{tabular}
    \caption{Statistics of \dataset dataset.}
    \label{tab:statistics}
\end{table*}

%% file: chapters_0115/4_experiment_settings/main.tex
\section{Experiment Settings}
To thoroughly understand \dataset and provide some promising directions for future ASTE research, we conduct experiments in multiple settings. This section will first provide an overview of the different experimental setups, then introduce the evaluation metric, and finally, describe the models employed in the experiments.
\input{chapters_0115/4_experiment_settings/set_up.tex}
\input{chapters_0115/4_experiment_settings/metric.tex}
\input{chapters_0115/4_experiment_settings/models.tex}

%% file: chapters_0115/4_experiment_settings/set_up.tex
\subsection{Setups}
We conduct a series of experiments under comprehensive training and testing setups:
\begin{itemize}
    \item \textit{In-domain}: we train and test the models with data from the same domain.
    
    \item  \textit{Single-source Cross-domain}: we train the models with data from a single-source domain and test them on a different target domain.

    \item \textit{Multi-source Cross-domain}: we train the models with data from multiple source domains and test them on a different target domain.

\end{itemize}
In the cross-domain setting, we regard Electronics, Fashion, Beauty, Home as the source domain, and Book, Pet, Toy, Grocery as the target domain.
More training details are shown in Appendix \ref{sec:app-exp-settings}.

%% file: chapters_0115/4_experiment_settings/metric.tex
\subsection{Evaluation Metric} 
Following \citet{xu-etal-2021-learning}, we employ the F1 score to measure the performance of different approaches. 
All the experimental results are reported using the average of 5 random runs.

%% file: chapters_0115/4_experiment_settings/models.tex

\subsection{Models}
This section presents the various approaches we evaluate on the \datasetwb. We first introduce representative models for ASTE and employ them as our baseline models. 
Then for the single-source cross-domain setting, we utilize current methods for ASTE and integrate some of them with the adversarial training \citep{ganin2016domain}, which is a widely-used technique in domain adaptation.

\paragraph{Baseline Models For ASTE.} 
We implement several representative models from various frameworks, including tagging-based, MRC-based, and generation-based frameworks.

\begin{itemize}
    \item Span-ASTE \citep{xu-etal-2021-learning}: a tagging-based method. It explicitly considers the span interaction between the aspect and opinion terms.
    \item BMRC \citep{chen2021bidirectional}: a MRC-based method. It  extracts aspect-oriented triplets and opinion-oriented triplets. Then it obtains the final results by merging the two directions. 
    \item  BART-ABSA \citep{yan-etal-2021-unified}: a generation-based method. It employs a pointer network and generates indices of the aspect term, opinion term, and sentiment polarity sequentially. 
    \item GAS \citep{zhang2021towards}: a generation-based method. It transforms the ASTE tasks into a text generation problem.
\end{itemize} 

\paragraph{Baseline Models For Single-source Cross-domain Setting.}
 We incorporate Span-ASTE and BMRC with adversarial training (AT), a common strategy in domain adaptation. Specifically, we apply a domain discriminator on different features for each method:
 
 \begin{itemize}
    \item BMRC+AT: We apply a domain discriminator on the token and [CLS] features. In this way, we can learn discriminative features by classifiers of the ASTE task and domain-invariant features by the domain discriminator induced by adversarial training.
    \item Span-ASTE+AT: 
    As the extraction in this method is based on the prediction of the span and pair representation, which are derived from token representation, we apply adversarial learning to the token representations, similar to the BMRC+AT model.
\end{itemize}

%% file: chapters_0115/5_results/main.tex
\input{tables/in_domain.tex}

\input{chapters/7_results_1225/fig-in-domain.tex}

\input{tables/main_results}

\section{Results}

This section presents thorough analyses of the challenges of the \dataset dataset and ASTE task, with the aim of better understanding these challenges and suggesting promising directions for future research. For each setting, we first show experimental results. Then we perform comprehensive analyses to investigate the challenges of the dataset and task, highlighting the limitations of current approaches. Finally, we provide promising directions for future research in this area.

\input{chapters_0115/5_results/in-domain.tex}

\input{chapters/6_experiment/fig_corr.tex}

\input{figures/fig_at_visualization}

\input{chapters_0115/5_results/single-source.tex}

\input{tables/all_multi_source.tex}

\input{chapters_0115/5_results/multi-source.tex}

%% file: tables/in_domain.tex
\begin{table*}
    \centering
    \small
    \begin{tabular}{cccccc}
    \toprule
        Method &  Electronics & Beauty & Fashion & Home & Average \\
    \midrule
        BMRC        &  41.95\scriptsize{$\pm$0.34} & 38.57\scriptsize{$\pm$0.97} & 44.87\scriptsize{$\pm$0.69} & 41.18\scriptsize{$\pm$0.66} & 41.64\\
        BART-ABSA   &  43.38\scriptsize{$\pm$1.37} & 41.13\scriptsize{$\pm$1.14} & 43.89\scriptsize{$\pm$0.82} & 40.56\scriptsize{$\pm$1.04} & 42.24\\
        GAS         &  47.10\scriptsize{$\pm$0.64} & 44.32\scriptsize{$\pm$0.52} & 47.80\scriptsize{$\pm$1.07} & 47.22\scriptsize{$\pm$1.13} & 46.61\\
        Span-ASTE   & \textbf{47.86}\scriptsize{$\pm$0.74} &\textbf{46.46}\scriptsize{$\pm$0.66} & \textbf{50.38}\scriptsize{$\pm$0.68} & \textbf{49.14}\scriptsize{$\pm$0.41} & \textbf{48.46}\\
    \bottomrule
    \end{tabular}
    \caption{F1 scores of in-domain ASTE on \datasetwb, and the best results  are highlighted in bold font. Span-ASTE is significantly better than other methods with p < 0.05.}
    \label{tab:in-domain}
\end{table*}

%% file: chapters/7_results_1225/fig-in-domain.tex
\begin{figure*}
     \centering
     \begin{subfigure}[b]{0.32\textwidth}
         \centering
         \includegraphics[width=\textwidth]{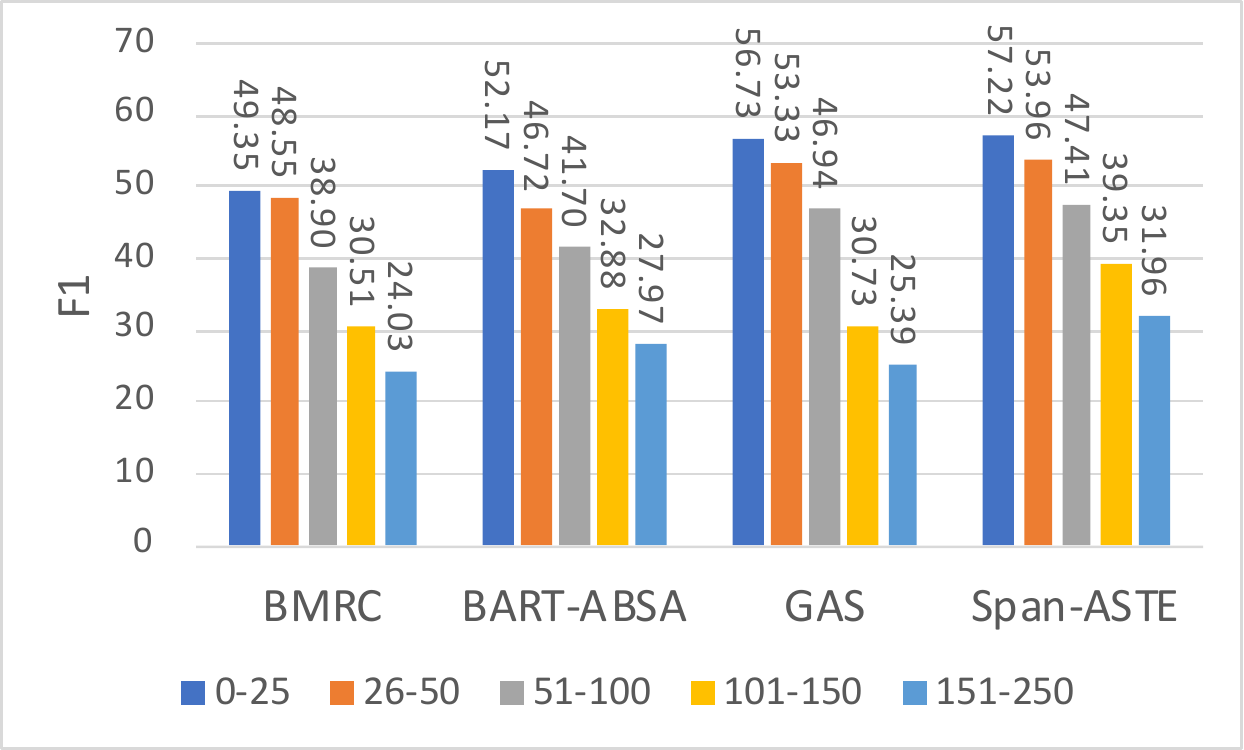}
         \caption{F1 scores on reviews of different lengths.}
         \label{fig:analyse-indomain-length}
     \end{subfigure}
        \hfill
     \begin{subfigure}[b]{0.32\textwidth}
         \centering
         \includegraphics[width=\textwidth]{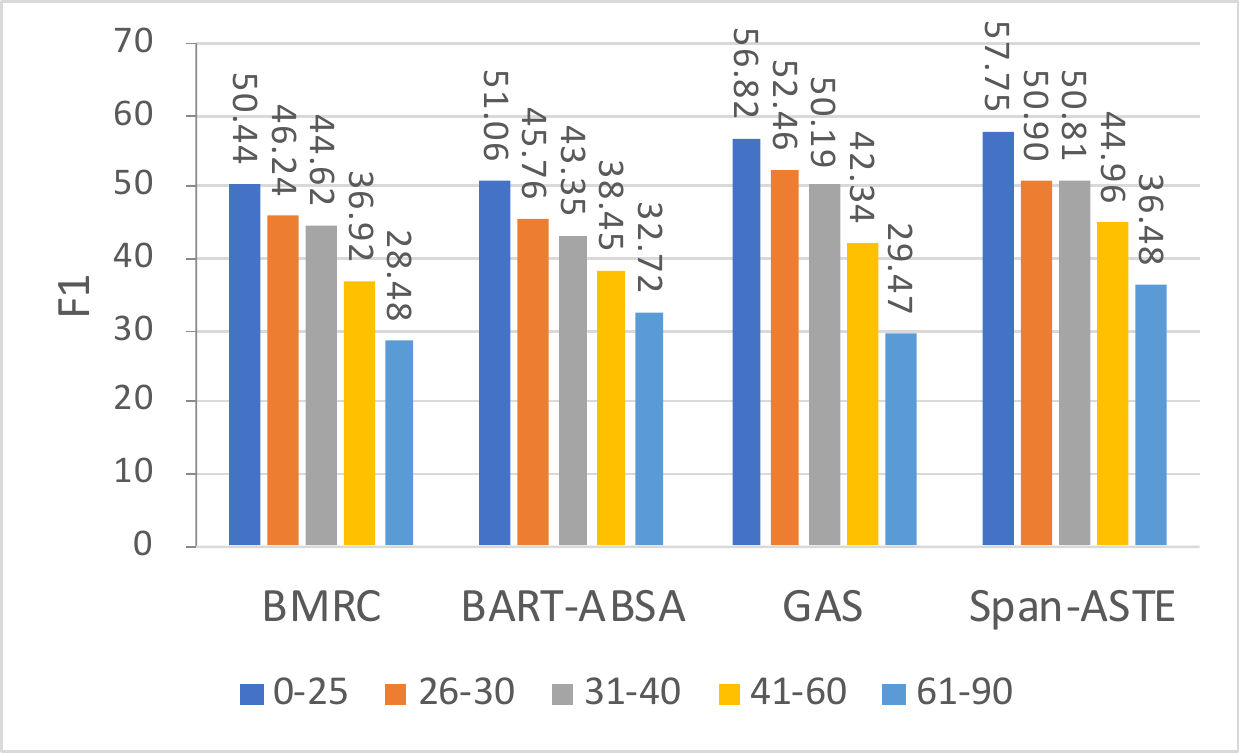}
         \caption{F1 scores on reviews of different numbers of POS 2-grams.}
         \label{fig:analyse-indomain-pos}
     \end{subfigure}
     \hfill
     \begin{subfigure}[b]{0.32\textwidth}
         \centering
         \includegraphics[width=\textwidth]{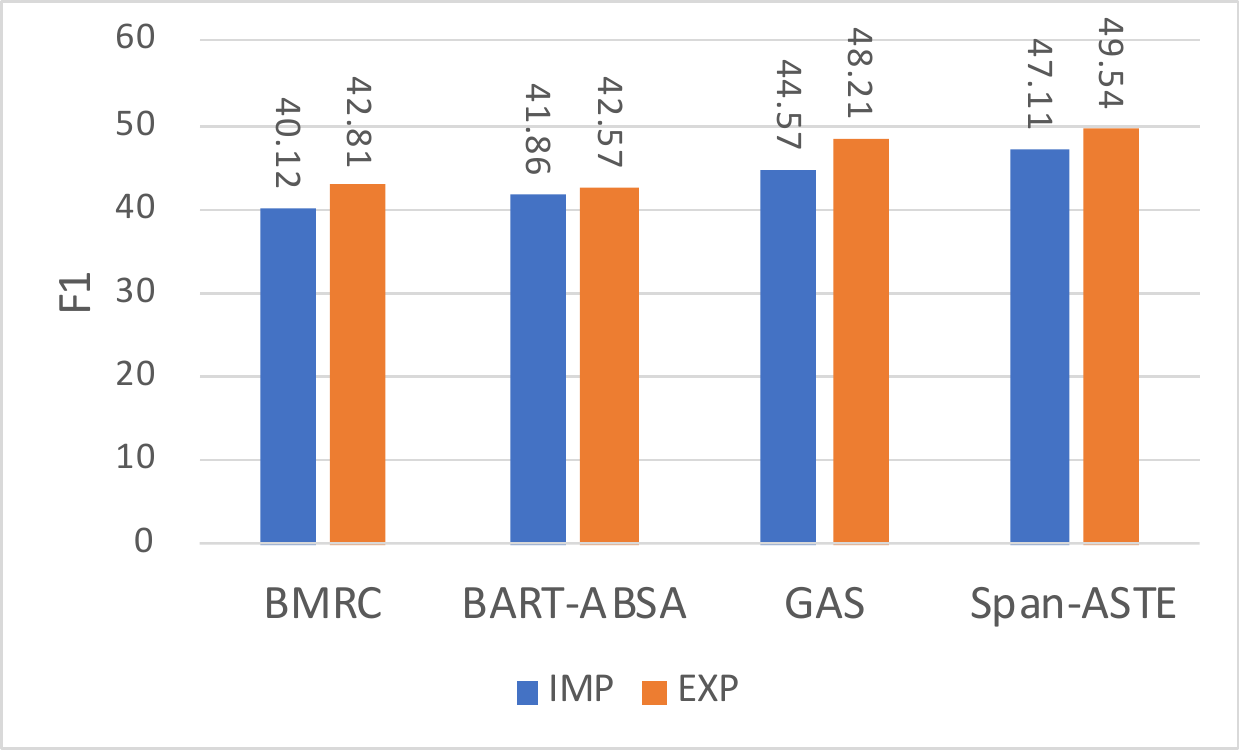}
         \caption{F1 scores on reviews of different aspect types.}
         \label{fig:analyse-indomain-imp}
     \end{subfigure}
        \caption{F1 scores on different review lengths, different numbers of POS 2-grams, and different aspect types. Results are average F1 scores on the four in-domain results. We can conclude that long reviews, complex sentences, and implicit aspect terms make \dataset more challenging. }
        \label{fig:analyse-indomain-ins}
\end{figure*}

%% file: tables/main_results.tex
\begin{table*}[h]
    \small
    \centering
    \newcolumntype{Z}{>{\centering\let\newline\\\arraybackslash\hspace{0pt}}X}
    \begin{tabular}{l c c c c c c c}
    
    \toprule
    Domain & BMRC &BART-ABSA & GAS& Span-ASTE & BMRC+AT & Span-ASTE+AT    \\
    \midrule
    Electronics$\to$ Book   & 33.74\scriptsize{$\pm$0.64} & 35.43\scriptsize{$\pm$0.61} & 35.57\scriptsize{$\pm$0.76} & \textbf{40.36}\scriptsize{$\pm$1.59} & 32.56\scriptsize{$\pm$1.57} & 39.58\scriptsize{$\pm$0.70}\\
    Beauty$\to$ Book        & 30.01\scriptsize{$\pm$1.29} & 30.24\scriptsize{$\pm$0.83} & 30.96\scriptsize{$\pm$0.99} & 38.58\scriptsize{$\pm$1.58} & 28.56\scriptsize{$\pm$0.88} & \textbf{38.79}\scriptsize{$\pm$0.89}\\
    Fashion$\to$ Book       & 31.71\scriptsize{$\pm$1.37} & 32.45\scriptsize{$\pm$1.77} & 36.26\scriptsize{$\pm$0.64} & \textbf{39.88}\scriptsize{$\pm$0.74} & 30.80\scriptsize{$\pm$1.59} & 39.60\scriptsize{$\pm$0.69}\\
    Home$\to$ Book          & 31.93\scriptsize{$\pm$1.79} & 33.48\scriptsize{$\pm$1.15} & 35.91\scriptsize{$\pm$0.75} & \textbf{39.45}\scriptsize{$\pm$0.73} & 31.31\scriptsize{$\pm$0.45} & 39.35\scriptsize{$\pm$0.93}\\
    \midrule 
    Electronics$\to$ Grocery& 39.68\scriptsize{$\pm$0.70} & 40.39\scriptsize{$\pm$1.28} & 39.16\scriptsize{$\pm$1.25} & \textbf{45.36}\scriptsize{$\pm$1.12} & 40.15\scriptsize{$\pm$0.55} & 43.90\scriptsize{$\pm$0.47}\\
    Beauty$\to$ Grocery     & 34.46\scriptsize{$\pm$0.85} & 34.22\scriptsize{$\pm$0.59} & 36.22\scriptsize{$\pm$0.50} & \textbf{40.32}\scriptsize{$\pm$0.81} & 34.91\scriptsize{$\pm$1.00} & 40.30\scriptsize{$\pm$1.64}\\
    Fashion$\to$ Grocery    & 37.18\scriptsize{$\pm$0.50} & 37.27\scriptsize{$\pm$1.08} & 40.13\scriptsize{$\pm$0.38} & \textbf{43.41}\scriptsize{$\pm$0.83} & 37.56\scriptsize{$\pm$0.29} & 42.29\scriptsize{$\pm$0.79}\\
    Home$\to$ Grocery       & 39.03\scriptsize{$\pm$1.78} & 38.56\scriptsize{$\pm$1.14} & 42.51\scriptsize{$\pm$0.37} & \textbf{43.74}\scriptsize{$\pm$1.02} & 38.83\scriptsize{$\pm$0.75} & 41.78\scriptsize{$\pm$1.79}\\
    \midrule 
    Electronics$\to$ Toy    & 42.39\scriptsize{$\pm$0.88} & 42.49\scriptsize{$\pm$1.13} & 43.55\scriptsize{$\pm$0.52} & 47.23\scriptsize{$\pm$0.38} & 41.38\scriptsize{$\pm$1.20} & \textbf{47.43}\scriptsize{$\pm$0.67}\\
    Beauty$\to$ Toy         & 35.66\scriptsize{$\pm$0.94} & 33.87\scriptsize{$\pm$1.05} & 37.95\scriptsize{$\pm$0.63} & \textbf{41.19}\scriptsize{$\pm$0.60} & 35.98\scriptsize{$\pm$1.14} & 40.86\scriptsize{$\pm$0.74}\\
    Fashion$\to$ Toy        & 41.13\scriptsize{$\pm$1.12} & 40.08\scriptsize{$\pm$1.15} & 42.78\scriptsize{$\pm$0.36} & \textbf{46.83}\scriptsize{$\pm$0.94} & 40.74\scriptsize{$\pm$1.20} & 46.09\scriptsize{$\pm$0.87}\\
    Home$\to$ Toy           & 40.26\scriptsize{$\pm$1.22} & 40.81\scriptsize{$\pm$1.27} & 44.16\scriptsize{$\pm$0.73} & \textbf{47.60}\scriptsize{$\pm$1.09} & 40.42\scriptsize{$\pm$0.51} & 46.47\scriptsize{$\pm$0.75}\\
    \midrule 
    Electronics$\to$ Pet    & 37.39\scriptsize{$\pm$0.60} & 36.88\scriptsize{$\pm$1.07} & 38.17\scriptsize{$\pm$0.89} & \textbf{41.04}\scriptsize{$\pm$0.48} & 36.70\scriptsize{$\pm$0.43} & 40.09\scriptsize{$\pm$0.77}\\
    Beauty$\to$ Pet         & 32.80\scriptsize{$\pm$1.07} & 32.07\scriptsize{$\pm$0.76} & 32.55\scriptsize{$\pm$0.57} & \textbf{36.41}\scriptsize{$\pm$0.40} & 32.76\scriptsize{$\pm$0.54} & 35.38\scriptsize{$\pm$0.83}\\
    Fashion$\to$ Pet        & 35.97\scriptsize{$\pm$0.84} & 34.92\scriptsize{$\pm$0.85} & 36.13\scriptsize{$\pm$0.74} & \textbf{40.57}\scriptsize{$\pm$0.57} & 36.07\scriptsize{$\pm$0.81} & 38.78\scriptsize{$\pm$0.81}\\
    Home$\to$ Pet           & 37.64\scriptsize{$\pm$1.03} & 37.26\scriptsize{$\pm$1.59} & 39.38\scriptsize{$\pm$0.70} & \textbf{41.42}\scriptsize{$\pm$0.51} & 37.24\scriptsize{$\pm$0.97} & 40.64\scriptsize{$\pm$1.05}\\
    \midrule
    Average     & 36.31 & 36.28 & 38.21 & \textbf{42.09} & 36.00 & 41.33\\
    \bottomrule
    \end{tabular}
    \caption{F1 scores of single-source cross-domain ASTE on \datasetwb. We highlight the best results in bold. Span-ASTE performs significantly better than BMRC, BART-ABSA, and GAS with p < 0.01.
    }
    \label{tab:main-results}
\end{table*}

%% file: chapters_0115/5_results/in-domain.tex
\subsection{In-domain}
\label{sec:app-in-domain}
 The overall in-domain experimental results are shown in Table \ref{tab:in-domain}. We first explore the limitations of baseline models to better understand challenges introduced by \datasetwb. Then we compare two representative models to find promising directions for future research. 

\paragraph{Challenges of \datasetwb.} Results of the in-domain experiments, reported in Table \ref{tab:in-domain}, show a performance drop compared with results on existing datasets (Span-ASTE gets a 59.38 score for the Laptop domain in \citet{xu-etal-2020-position} and 47.86 for the Electronics domain in \datasetwb). To further investigate the challenges of \datasetwb, we further analyze the performance under different review lengths, sentence complexity, and aspect types.

\begin{itemize}
    \item \textbf{Length. } Figure \ref{fig:analyse-indomain-length} illustrates the relationship between the review length and the model performance on \datasetwb. Results show that the performance of the models decreases as the length of the review increases. This indicates long reviews present a significant challenge for current models.
    \item \textbf{Sentence Complexity. } We quantify the sentence complexity by calculating the number of 2-grams in the part-of-speech (POS) sequence of the review text. Then we analyze  the relationship between the extraction performance and sentence complexity. As shown in Figure \ref{fig:analyse-indomain-pos}, we observe a decline in performance with an increase in sentence complexity. This highlights the challenges posed by the diversified expression presented in \datasetwb.
    \item \textbf{Aspect Types. } In Figure \ref{fig:analyse-indomain-imp}, we analyze the performance of models on triplets with implicit and explicit aspect terms. Results demonstrate that the extraction of implicit aspect terms is more challenging  than that of explicit aspect terms. The inclusion of both implicit and explicit aspect terms in \dataset makes it more challenging for ASTE.  
\end{itemize}
We can conclude that long reviews, complex sentences and implicit aspect terms make \dataset a challenging dataset.

\paragraph{Model Comparison.}
Prior works have demonstrated that GAS \citep{zhang2021towards} outperforms Span-ASTE \citep{xu-etal-2021-learning} on the dataset of \citet{xu-etal-2020-position}. 
However, as shown in Table \ref{tab:in-domain}, Span-ASTE outperforms GAS on \datasetwb. We conduct further analysis and discover that the reversal in performance can be attributed to the long instance lengths and complex sentence patterns presented in \datasetwb. As illustrated in Figure \ref{fig:analyse-indomain-ins}, GAS achieves comparable results when the reviews are short and simple, but its performance declines more sharply when encountering long and complex reviews in comparison to Span-ASTE. This can be attributed to the generation-based nature of GAS, which is prone to forgetting the original text and misspellings or  generating phrases that do not appear in the review when dealing with long reviews. In contrast, the tagging-based nature of Span-ASTE only identifies the start and end tokens for the aspect and opinion terms, which is less affected by the complexity and length of the reviews. This analysis provides insights for future generation-based methods: \textit{implementing a comparison algorithm between the generated output and the original input text and making modifications if they are mismatched.}

%% file: chapters/6_experiment/fig_corr.tex

\begin{figure*}
     \centering
     \begin{subfigure}[b]{0.4\textwidth}
         \centering
         \includegraphics[width=\textwidth]{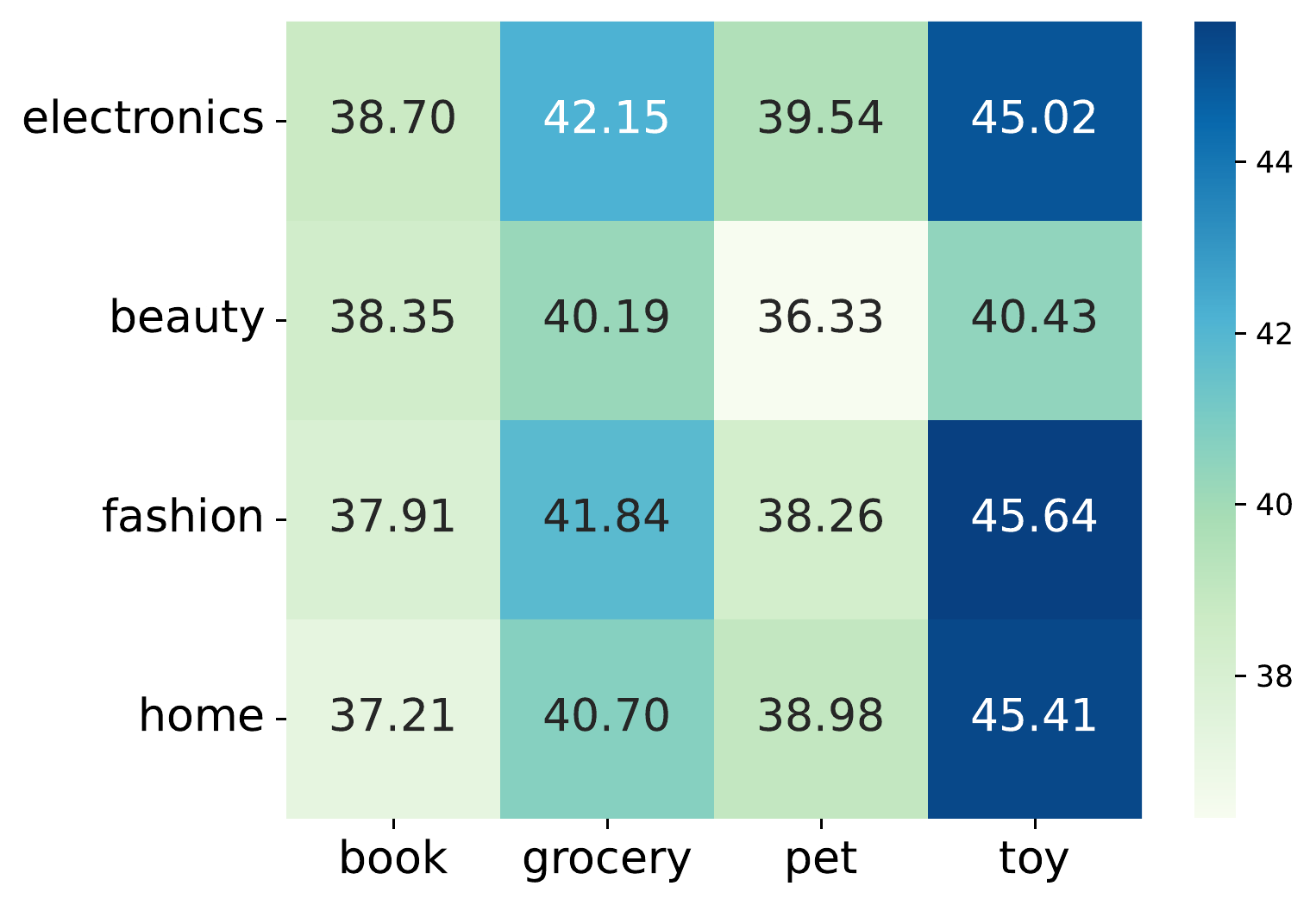}
         \caption{Model performances on single-source cross-domain ASTE.}
         \label{fig:corr-baseline}
     \end{subfigure}\quad
     \begin{subfigure}[b]{0.4\textwidth}
         \centering
         \includegraphics[width=\textwidth]{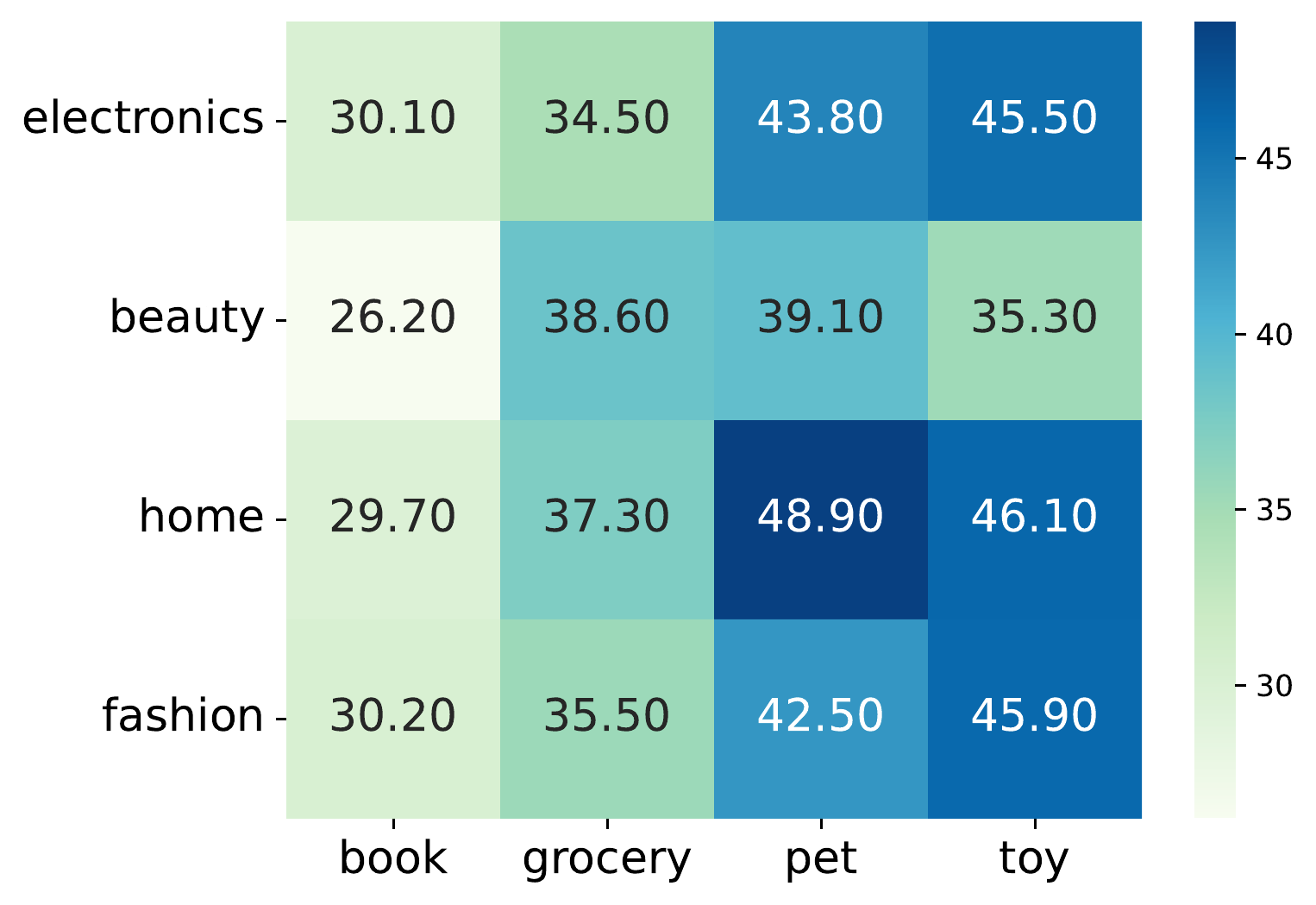}
         \caption{Vocabulary overlaps (\%) between the source domains and target domains.}
         \label{fig:corr-similarity}
     \end{subfigure}
        \caption{Correlation analysis between the model performance and domain similarity. }
        \label{fig:corr}
\end{figure*}

%% file: figures/fig_at_visualization.tex
\begin{figure*}[h]
     \centering
     \begin{subfigure}[b]{0.45\textwidth}
         \centering
         \includegraphics[width=\textwidth]{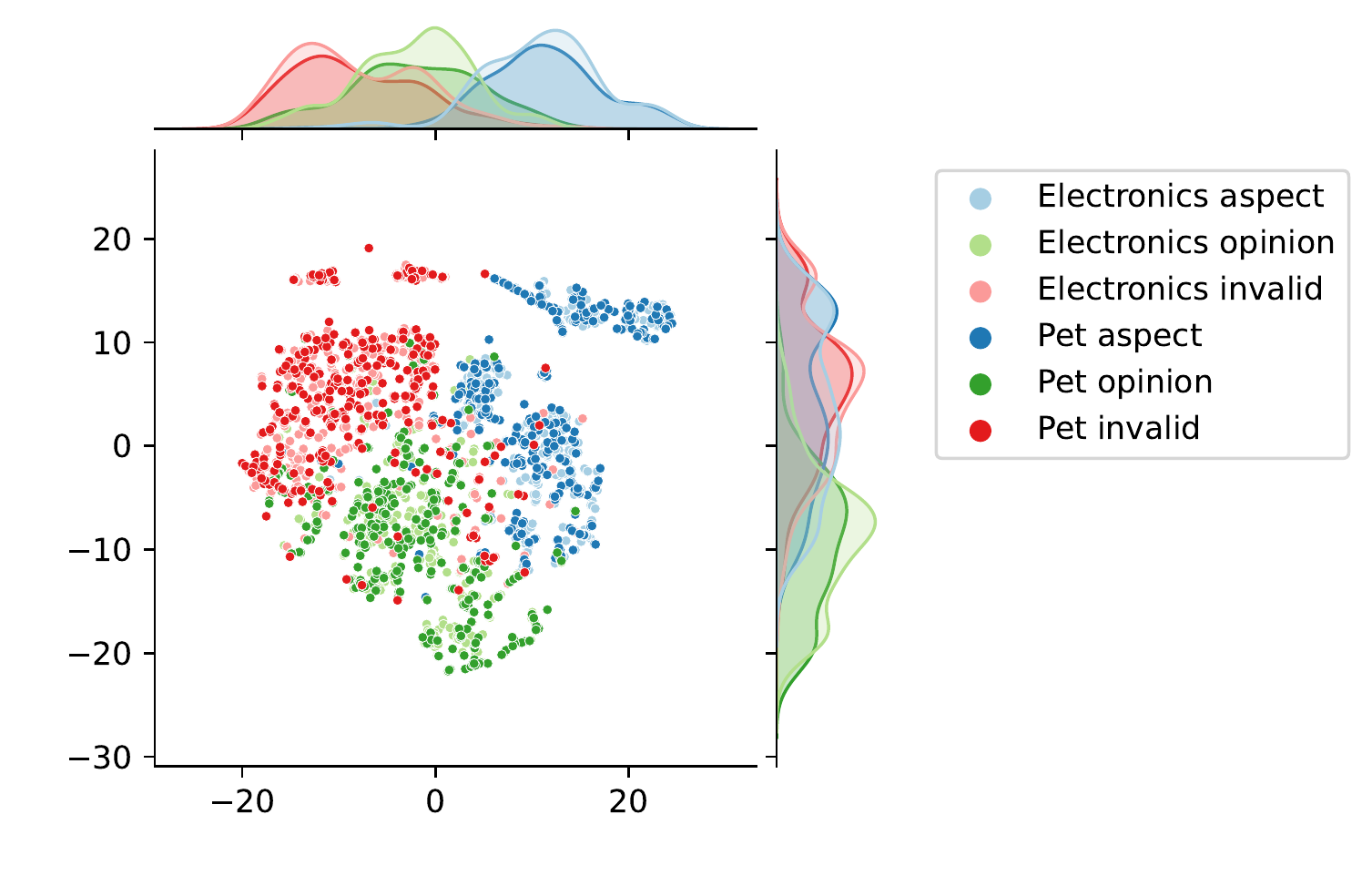}
         \caption{Span-ASTE}
         \label{fig:tsne-span}
     \end{subfigure}
     \begin{subfigure}[b]{0.45\textwidth}
         \centering
         \includegraphics[width=\textwidth]{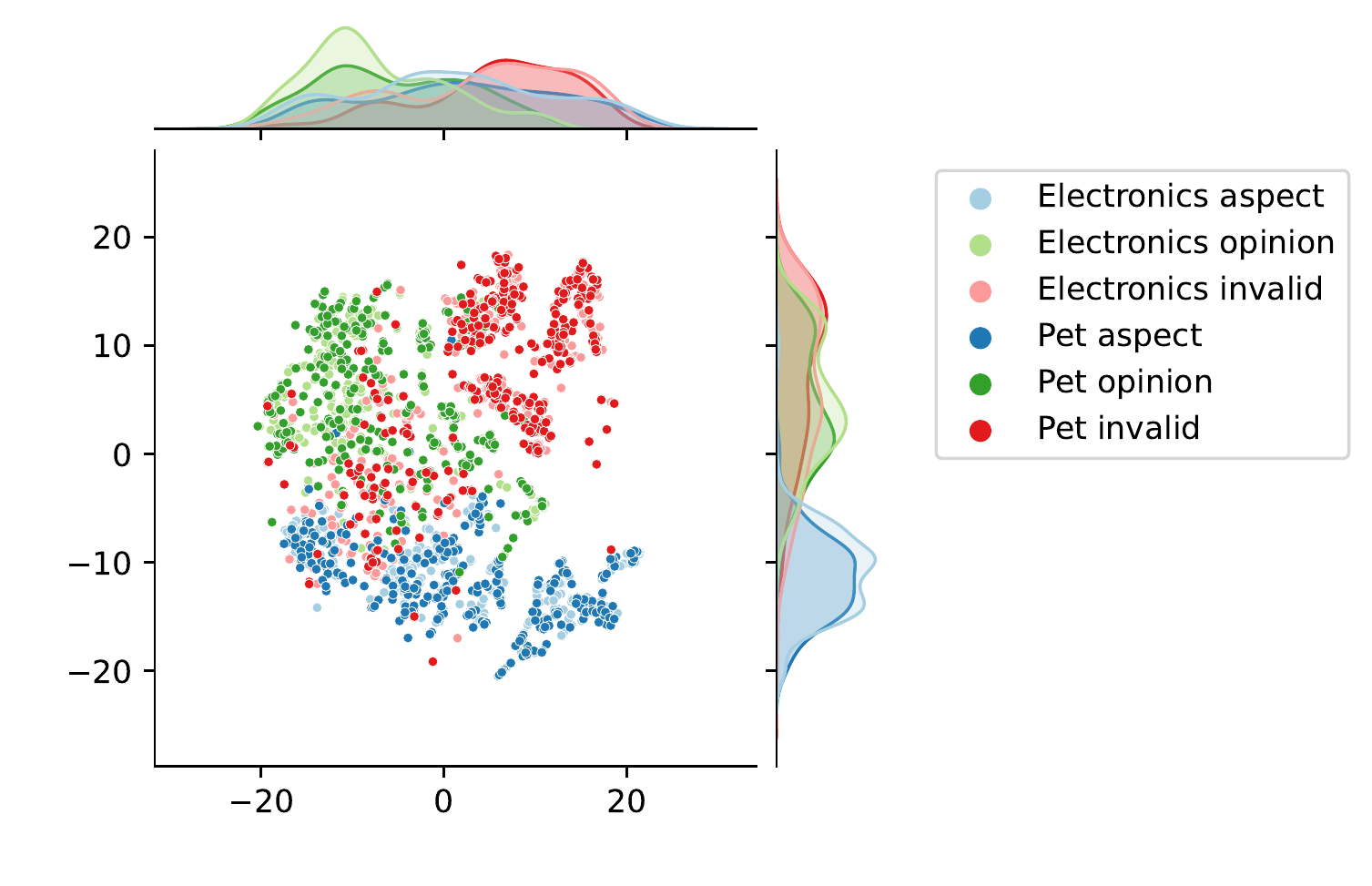}
         \caption{Span-ASTE+AT}
         \label{fig:tsne-at}
     \end{subfigure}
        \caption{T-SNE visualization of Span-ASTE and Span-ASTE+AT on Electonics$\to$Pet. 
        Compared with Span-ASTE, the features of different categories in Span-ASTE+AT are less discriminable, especially on the x-axis.
        }
        \label{fig:at-visualization}
\end{figure*}

%% file: chapters_0115/5_results/single-source.tex
\subsection{Single-source Cross-domain}\label{sec:exp-single}
We present the overall single-source cross-domain experimental results in Table \ref{tab:main-results} (additional training details of adversarial training are shown in Appendix \ref{sec:app-adversarial}). We then conduct a correlation analysis to investigate the factors that impact the transfer performance. Additionally, we analyze current domain adaptation strategies and provide insights for future research in this area.



 \paragraph{Model Performance v.s. Domain Similarity.} 
 Table \ref{tab:main-results} reveals that performance varies significantly when transferring from different source domains. For instance, the F1 score of Span-ASTE on Home$\to$ Toy is 47.60, which is 6.41 higher than that on Beauty$\to$Toy. To gain insights into the factors that impact the transfer performance, we conduct the Pearson Correlation Analysis \citep{benesty2009pearson} on the relationship between the model performance and domain similarity based on Span-ASTE. Specifically, we fix the number of training data for each source domain to alleviate the impact of data volume. Following \citet{liu2021crossner}, we measure the domain similarity by computing vocabulary overlaps, using the top 1k most frequent words in each domain excluding stopwords. 
 Results in Figure \ref{fig:corr} show that there is a positive correlation (with a 0.52 Pearson correlation coefficient) between model performance and domain similarity. 
 This indicates that large domain discrepancy is a huge challenge for the single-source cross-domain ASTE task. Therefore, \textit{reducing the domain discrepancy is a promising way to improve the transfer performance in cross-domain ASTE.}

\paragraph{With AT v.s. Without AT.} 
We compare the performance of BMRC and BMRC+AT, as well as Span-ASTE and Span-ASTE+AT in Table \ref{tab:main-results}. Results indicate that adversarial training has a negative impact on the performance of cross-domain ASTE. To investigate the cause of performance degradation, we visualize the representations learned from the models when transferring from the Electronics to the Pet domain in Figure \ref{fig:at-visualization}. 
Compared with Span-ASTE, the features of different categories in Span-ASTE+AT are less discriminable, especially on the x-axis.
We attribute the performance drop to feature collapse \citep{tang2020discriminative} induced by adversarial training. This occurs when the model focuses on learning domain-invariant features while ignoring the discriminability of each category. This issue is particularly pronounced in the ASTE task, as it requires fine-grained discrimination for three factors. 
Future research in cross-domain ASTE could focus on \textit{developing methods that can learn domain-invariant features while maintaining the discriminability of each category.}

%% file: tables/all_multi_source.tex
\begin{table*}[]
    \small
    \centering
    \begin{tabular}{lccccc}
    \toprule 
    Domain     & Book & Grocery & Toy & Pet & Average  \\
    \midrule
    B + E       & 40.94\scriptsize{$\pm$1.14} & 45.94\scriptsize{$\pm$0.56}& 48.07\scriptsize{$\pm$0.75} & 41.70\scriptsize{$\pm$0.79} & 44.16\\
    E + F       & 40.59\scriptsize{$\pm$0.86} & 46.14\scriptsize{$\pm$0.98} & 50.10\scriptsize{$\pm$0.83} & 41.90\scriptsize{$\pm$0.79} & 44.68\\
    E + H       & 41.08\scriptsize{$\pm$1.50} & 45.34\scriptsize{$\pm$0.50} & 48.97\scriptsize{$\pm$0.69} & 42.70\scriptsize{$\pm$0.41} & 44.52\\
    B + F       & 41.13\scriptsize{$\pm$0.51} & 44.60\scriptsize{$\pm$0.73} & 46.81\scriptsize{$\pm$0.91} & 41.03\scriptsize{$\pm$0.59} & 43.39\\
    B + H       & 41.31\scriptsize{$\pm$1.18} & 44.31\scriptsize{$\pm$0.84} & 47.12\scriptsize{$\pm$0.87} & 41.67\scriptsize{$\pm$0.93} & 43.60\\
    F + H       & 42.40\scriptsize{$\pm$0.31} & 45.97\scriptsize{$\pm$0.56} & 49.11\scriptsize{$\pm$0.74} & 43.12\scriptsize{$\pm$0.70} & 45.15\\
    B + E + F   & 41.34\scriptsize{$\pm$0.73} & 46.28\scriptsize{$\pm$0.78} & 49.44\scriptsize{$\pm$1.08} & 41.12\scriptsize{$\pm$0.58} & 44.80\\
    B + E + H   & 41.21\scriptsize{$\pm$0.48} & 45.39\scriptsize{$\pm$0.49} & 49.33\scriptsize{$\pm$0.32} & 43.48\scriptsize{$\pm$1.12} & 44.85\\
    E + F + H   & 41.44\scriptsize{$\pm$0.71} & 45.39\scriptsize{$\pm$0.80} & 50.24\scriptsize{$\pm$0.57} & 43.59\scriptsize{$\pm$1.06} & 45.17\\
    B + F + H   & 41.73\scriptsize{$\pm$0.38} & 45.40\scriptsize{$\pm$0.63} & 48.55\scriptsize{$\pm$0.71} & 43.46\scriptsize{$\pm$0.19} & 44.79\\
    ALL         & 41.83\scriptsize{$\pm$0.57} & 46.07\scriptsize{$\pm$0.47} & 50.16\scriptsize{$\pm$0.49} & 43.62\scriptsize{$\pm$1.32} & 45.42\\
    \bottomrule
    \end{tabular}
    \caption{Comparison results of multi-source cross-domain ASTE on Span-ASTE. We use abbreviations due to space limitation: B: Beauty, H: Home, F: Fashion, E: electronics, ALL: all the source domains. }
    \label{tab:app-multi-source}
\end{table*}

%% file: chapters_0115/5_results/multi-source.tex
\subsection{Multi-source Cross-domain}
In this section, we conduct multi-source cross-domain experiments with the number of source domains varying from 2 to 4. The results are shown in Table \ref{tab:app-multi-source}. Our results reveal instances of negative transfer. For example, F + H $\to$ Book > E + F + H $\to$ Book.

Negative transfer indicates that transferring from some domains could harm the learning of the target domain \citep{guo-etal-2018-multi}. To further investigate the phenomenon of negative transfer, we conduct an analysis on the relationship between domain similarity and transfer performance. We further compare the results with the domain similarity in Figure \ref{fig:corr-similarity}. The results indicate that half of the negative transfers are observed when transferring from the source domains with the least similarity to the target domain (e.g., F + H $\to$ Toy > B  + F + H $\to$ Toy. Adding the Beauty domain leads to negative transfer and it is the least similar domain with Toy). This suggests that domain similarity can serve as a useful guideline for selecting source domains in future multi-source cross-domain research.

%% file: chapters_0115/6_conclusion.tex
\section{Conclusion and Future Directions}
We propose \datasetwb, a manually-annotated ASTE dataset collected from Amazon. Compared with existing ASTE datasets, \dataset contains reviews of various lengths, diverse expressions, more aspect types and covers more domains, which indicates \dataset is a suitable testbed to verify the ability of ASTE methods in real-world scenarios.
We explore the dataset in multiple scenarios, i.e., in-domain, single-source cross-domain, and multi-source cross-domain and provide some promising directions for future research. For the in-domain setting, we compare the results between existing datasets and \dataset and find that the long reviews, complex sentences, and implicit aspect terms make \dataset a challenging dataset. For the single-source cross-domain setting, 
we observe that domain similarity and cross-domain performance are positively correlated. 
Furthermore, analysis of adversarial training shows that simply learning domain-invariant features may lead to feature collapse and result in the loss of task-specific knowledge. Therefore, it is important to design appropriate methods to reduce the domain discrepancy while preserving fine-grained task features for ASTE tasks.
In multi-source domain adaptation, we find that most of the negative transfer comes from dissimilar source-target pairs, pointing out that domain similarity can be a domain selection guideline for future research.
In conclusion, we hope that our dataset \dataset and analyses will contribute to the promotion of ASTE research.

%% file: chapters_0115/7_limitations.tex
\section*{Limitations}
We analyze the limitations of this study from the following perspectives:
\begin{itemize}
    \item The ASTE task extracts the sentiment triplets from a review, while the Aspect Sentiment Quad Prediction (ASQP) task adds an aspect category based on the triplets and provides more comprehensive information. Defining the aspect category for each domain is also hard work.
    Future work can take the aspect category into consideration.
    \item All the models are evaluated by F1 score, in which only exact matching can be considered correct. This metric can not differentiate between partially matching and completely mismatching and is not the best choice for a challenging dataset like \datasetwb. Future work can include some partially matching metrics for this task.
    \item There is no specifically designed method for cross-domain ASTE. But we analyze the challenges of this task in detail. We are planning to design a new method for cross-domain ASTE based on the analysis results.
\end{itemize}

%% file: chapters_0515/8_acknowledgement.tex
\section*{Acknowledgments}
This work is done during Ting Xu's internship at ByteDance. We would like to thank the anonymous reviewers for their insightful comments. Zhen Wu is the corresponding author. Ting Xu would like to thank Siyu Long for his constructive suggestions. This work is supported by National Natural Science Foundation of China (No. 62206126, 61936012, and 61976114).

%% file: chapters_0115/8_appendix/main.tex
\clearpage
\label{sec:appendix}
\input{chapters_0115/8_appendix/dataset.tex}
\input{chapters_0115/8_appendix/labeling_guideline.tex}

\input{tables/app_case.tex}

\section{Experiments}
\input{chapters_0115/8_appendix/experiment_setting.tex}
\input{chapters_0115/8_appendix/multi.tex}
\input{chapters_0115/8_appendix/adversarial_exp.tex}

%% file: chapters_0115/8_appendix/dataset.tex
\section{Dataset}
\label{sec:app-dataset}
\subsection{Dataset Details}
\paragraph{Domain Selection Process.} First, we select the most popular four domains based on the reports of Top selling products on \href{https://www.junglescout.com/blog/amazon-product-categories/}{Amazon} and \href{https://agenciaeplus.com.br/en/produtos-mais-vendidos-na-internet-2022-veja-a-lista-dos-10/}{the Internet}. For these four domains, we annotate more data to enable research on ASTE with more realistic and representative reviews. Then, to enable more comprehensive research like the cross-domain setting, we randomly select another four domains and annotate fewer data for testing in the cross-domain setting. 
\paragraph{Sampled Reviews.} We sample two reviews for each domain and demonstrate them in Table \ref{tab:app-case}. The first column displays the review text. And the second column shows the extracted triplets (aspect terms, opinion terms, sentiment polarity).
\paragraph{Annotator Compensation.} As described in Section \ref{sec:annotation}, we hire 18 workers in the annotation and follow strict quality control to ensure the quality of the annotation. We ensure the privacy right of workers is respected in the annotation process.
All workers have been paid above the local minimum wage and agree to use the dataset for research purposes.
\paragraph{License.}\dataset will be publicly avaliable under the terms of \href{https://www.wikidata.org/wiki/Q42553662}{CC BY-NC-SA 4.0 License}. The dataset is for academic use, which is consistent with its origination Amazon \citep{ni-etal-2019-justifying} dataset.

%% file: chapters_0115/8_appendix/labeling_guideline.tex
\subsection{Annotation Guidelines}
\label{sec:app-guideline}
The purpose of this annotation guideline is to provide a consistent and accurate method for annotating reviews. The task of the annotator is to identify triplets comprising of the following types: 
\begin{itemize}
    \item \textbf{Aspect Term. } An Aspect term refers to a part or an attribute of products or services. Sometimes, the aspect term may not appear explicitly in the instance, i.e., implicit aspect \citep{poria2014rule}. We keep the triplets with implicit aspects.
    \item \textbf{Opinion Term. } An opinion term is a word or phrase that expresses opinions or attitudes towards aspects.
    \item \textbf{Sentiment Polarity. } Sentiment polarity is the sentiment type of the opinion term, which is divided into three categories: positive, negative, and neutral.
\end{itemize}

\paragraph{Annotation Guidelines for Aspect Terms.} Aspect terms can be divided into three types: (1) a part of the product, (2) an attribute of the product or store, (3) an attribute of a part of the product. Note, if the review expresses opinions toward some targets which is not explicitly mentioned, we annotate them as implicit aspect (NULL). If we have two aspect terms split with "and", ",", etc, label them separately. \textit{Examples of aspect terms: battery, price, outlook, taste, customer support.}

\paragraph{Annotation Guidelines for Opinion Terms.} Opinion terms usually expressed opinions or attitudes. (1) Label as much information as possible about "opinion of the product, user experience, an emotion about buying the product", etc. (2) The information selected should not change the original semantics. (3) Opinion terms are usually adjectives/adverbs, and they can be a phrase or a single word. (4) "definitely" or "very" alone are not enough to be labeled as an opinion term. (5) If we have two opinion terms split with "and", ",", etc, label them separately. \textit{Examples of opinion terms: so good, expensive, love, unfriendly.}

\paragraph{Annotation Guidelines for Sentiment Polarities.} Sentiment polarities can be divided into three types: (1) positive: the review expresses a positive attitude toward a specific aspect, (2) negative: the review expresses a negative attitude toward a specific aspect, (3) neutral: the review expresses no obvious positive or negative attitudes but it expresses an attitude. \textit{Examples of sentiment polarities: (price, expensive, negative), (battery life, long, positive), (outlook, just okay, neutral).}

\paragraph{Abandon Cases.} If the review is just some meaningless word, e.g. hhhh, ahhhh, oooooo, then abandon it.

\paragraph{Annotations Steps.}
When annotating the reviews, please follow the steps below:
\begin{enumerate}
    \item Start annotating the reviews from left to right one by one using appropriate order from aspect terms, opinion terms to sentiment polarities.
    \item Please make sure to select the right text boundary for aspect terms and opinion terms before clicking the "Mark this" button.
    \item Please make sure to select a suitable category for sentiment polarities.
    \item After double-checking the current triplet, click the "complete" button to proceed to the next triplet.
    \item If the review contains more than one triplet, go through the same steps to annotate them and click "Submit" to proceed to the next review.
    \item If the review satisfies the condition of abandon case, click "abandon" and proceed to the next review.
\end{enumerate}

\paragraph{Appeal Process.}
During the check phase, the verifiers are not always right. Specifically, we introduce an Appeal process to ensure the high quality of dataset in the check phase. The detailed process is as follows:
\begin{enumerate}
    \item If the verifier agrees with the annotation of the annotator, the annotation is added to the dataset. Otherwise, the verifier will annotate the data and send it to the annotator as feedback. If the annotator accepts the feedback, the annotation from the verifier is added to the dataset.
    \item If the annotator disagrees with the feedback of the verifier, the annotator will appeal. The corresponding data will be discussed by all verifiers and annotators until they reach an agreement. After that, the new annotation is added to the dataset. 
\end{enumerate}

From the above process, verifiers sometimes play a role in annotators and they work together to ensure the high quality of dataset. We will add the above details in the revision.

%% file: tables/app_case.tex
\begin{table*}[]
    \small
    \centering
    \begin{tabularx}{\textwidth}{XX}
    \toprule
    \textbf{Electronics}\\
    \bottomrule
    Good case . It has minimum padding that makes the phone feel secure but too much to be burdensome . The card slots are convenient . &(card slots, convenient, POS); (case, Good, POS); (padding, minimum, POS).\\
    \midrule
    They work very well . The connections are nice and tight and I loose no quality over the length of the cord .&(connections, nice, POS); (connections, tight, POS); (length of the cord, loose no quality over, POS); (work, very well, POS).\\
     \bottomrule
    \textbf{Beauty}\\
    \bottomrule
     This smells so good exactly like the body lotion and spray mist . It is a soft femine fragrance not too overpowering . I would highly recommend ! & (NULL, would highly recommend, POS); (fragrance, not too overpowering, POS); (fragrance, soft femine, POS); (smells, so good, POS).\\
    \midrule
     Revision products are awesome . I switch away from this in the winter months since it s not as moisturizing as an emollient based cleanser . &(Revision, awesome, POS); (NULL, not as moisturizing, NEG).\\
    \bottomrule
    \textbf{Home}\\
    \bottomrule
    I think i just started a new hobby , this Victorinox is pretty awesome . I might have to get a few more designs .&(Victorinox, pretty awesome, POS).\\
    \midrule
    Too big and cumbersome to be useful . If design were downsized a couple of inches , they would be perfect . Much too big for my salad bowls . & (NULL, Much too big, NEG); (NULL, Too big, NEG); (NULL, cumbersome, NEG).\\
    \bottomrule
    \textbf{Fashion}\\
    \bottomrule
    the fabric is very , very thin . It should be underwear , not a regular shirt . fits too snug and is see through &(NULL, see through, NEG); (fabric, very thin, NEG); (fits, too snug, NEG).\\
    \midrule
    What can I say that you don t already know ? Classic Chucks . One of the best , most versatile , and most durable sneakers out there . &(Chucks, Classic, POS); (sneakers, One of the best, POS); (sneakers, most durable, POS); (sneakers, most versatile, POS).\\
    \bottomrule
    \textbf{Book}\\
    \bottomrule
    It was a good book , haven t read it before bow . Very interesting , with good suspense and humor mixed in ! 10 / 10 &(NULL, Very interesting, POS); (book, good, POS); (humor, good, POS); (suspense, good, POS).\\
    \midrule
    Challenging to read but worth it . Persevere to the end of the book . Ask God to open your mind and heart . &(NULL, Challenging to read, NEU); (NULL, worth, POS).\\
    \bottomrule
    \textbf{Grocery}\\
    \bottomrule
    These are the best Slim Jims every , I have tried them , but the taste , the texture of the Honey BBQ is the Greatest .&(Slim Jims, best, POS); (taste, Greatest, POS); (texture, Greatest, POS).\\
    \midrule
    A good almond extract , and I like that it s organic . I took off one star for high price . Should not cost that much .&(almond extract, good, POS); (organic, like, POS); (price, high, NEG).\\
    \bottomrule
    \textbf{Pet}\\
    \bottomrule
    This is a good prefilter . The clear plastic adapters are a little brittle so be careful when attaching to your filter input . &(clear plastic adapters, a little brittle, NEG); (prefilter, good, POS).\\
    \midrule
    Bought these as cat treats . Have to break them up , but my cat goes crazy for them , and this tub is lasting forever . &(NULL, goes crazy for, POS); (tub, lasting forever, POS).\\
    \bottomrule
    \textbf{Toy}\\
    \bottomrule
    Super simple dynamics , and a great game for friends , family , kids , etc . ! Easy to learn , and variety of play is good . &(NULL, Easy to learn, POS); (dynamics, Super, POS); (dynamics, simple, POS); (game, great, POS); (variety of play, good, POS).\\
    \midrule
    My 1 year old loved this for his birthday . Such a fun , easy toy . I would buy this again and again . &(NULL, loved, POS); (toy, easy, POS); (toy, fun, POS).\\
    \bottomrule
    \end{tabularx}
    \caption{Sampled reviews for each domain in \datasetwb.}
    \label{tab:app-case}
\end{table*}

%% file: chapters_0115/8_appendix/experiment_setting.tex
\subsection{Training Details}
\label{sec:app-exp-settings}
We utilize the pretrained model provided by \href{https://huggingface.co/}{HuaggingFace} and run all the experiments on NVIDIA A100 GPU with pytorch.
For the hyper-parameters of the baseline models, we follow the original settings in their paper \citep{chen2021bidirectional, yan-etal-2021-unified, zhang-etal-2021-towards, xu-etal-2021-learning}. For adversarial training, we follow the implementation in \citet{ganin2016domain}. One hyper-parameter in this method is $\alpha$, the ratio of training the generator to the discriminator. We search $\alpha$ in $\{1, 3, 5, 7, 10, 15, 20, 30, 50, 100\}$ and $\{1, 10, 30, 50, 100, 500, 700, 800, 1000, 1500\}$ for Span-ASTE + AT and BMRC + AT, respectively. For each value, we conduct experiments with 5 random seeds and set $\alpha$ by the F1 score on the development set. We set $\alpha=10$ for Span-ASTE+AT and $\alpha=800$ for BMRC+AT. The parameter search costs about 1000 GPU hours.

%% file: chapters_0115/8_appendix/multi.tex

%% file: chapters_0115/8_appendix/adversarial_exp.tex
\subsection{Detailed Results For Adversarial Training}
\label{sec:app-adversarial}
We search the hyper-parameter $\alpha$ for adversarial learning on the development set.
Detailed experiment results are shown in Table \ref{tab:span-at} and Table \ref{tab:bmrc-at}. We can observe that adversarial training is parameter-sensitive. 
\input{tables/span_at.tex}
\input{tables/bmrc_at.tex}

%% file: tables/span_at.tex
\begin{table*}[]
    \small
    \centering
    \begin{tabular}{cccccccccccccccc}
    \toprule
         $\alpha$ & 1 & 3 & 5 & 7 & 10 & 15 & 20 & 30 & 50 & 100  \\
    \midrule
        E$\to$K & 33.79 & 37.46 & 38.36 & 38.74 & 38.74 & 39.46 & 35.68 & 38.99 & 35.17 & 35.70\\
        B$\to$K & 31.55 & 35.20 & 37.95 & 36.63 & 37.83 & 36.77 & 35.37 & 34.99 & 36.47 & 28.41\\
        F$\to$K & 32.15 & 39.83 & 39.74 & 39.93 & 39.95 & 39.38 & 38.88 & 39.86 & 40.13 & 38.93 \\
        H$\to$K & 35.15 & 35.64 & 40.12 & 38.12 & 39.96 & 38.10 & 38.45 & 37.40 & 32.99 & 38.49\\
    \midrule
        E$\to$G & 39.86 & 44.71 & 44.86 & 45.11 & 44.39 & 45.66 & 44.23 & 44.59 & 44.46 & 45.48\\
        B$\to$G & 18.92 & 40.90 & 40.69 & 42.16 & 41.01 & 40.39 & 40.49 & 41.41 & 40.51 & 41.49\\
        F$\to$G & 38.23 & 43.98 & 43.36 & 43.88 & 44.89 & 42.65 & 44.40 & 44.23 & 43.97 & 43.78\\
        H$\to$G & 27.58 & 44.18 & 45.00 & 43.74 & 44.38 & 44.46 & 44.25 & 44.15 & 43.52 & 44.12\\
    \midrule
        E$\to$T & 33.21 & 47.82 & 48.65 & 48.48 & 47.23 & 47.95 & 47.57 & 48.26 & 48.64 & 47.33\\
        B$\to$T & 30.05 & 41.54 & 41.12 & 43.13 & 42.65 & 42.03 & 41.76 & 41.35 & 41.52 & 42.45\\
        F$\to$T & 33.63 & 46.47 & 47.12 & 47.65 & 47.16 & 47.73 & 48.00 & 47.31 & 46.53 & 47.08\\
        H$\to$T & 42.09 & 48.27 & 48.24 & 48.30 & 48.44 & 48.93 & 46.19 & 47.01 & 48.41 & 48.91\\
    \midrule
        E$\to$P & 33.51 & 41.18 & 41.95 & 41.19 & 41.43 & 41.64 & 41.81 & 42.01 & 42.36 & 42.06\\
        B$\to$P & 18.03 & 35.96 & 37.05 & 35.71 & 36.92 & 36.08 & 36.26 & 36.72 & 37.15 & 36.10\\
        F$\to$P & 35.44 & 39.30 & 39.87 & 39.98 & 40.00 & 40.07 & 40.24 & 39.61 & 40.41 & 40.40\\
        H$\to$P & 35.41 & 41.28 & 41.55 & 42.24 & 41.35 & 42.51 & 41.41 & 37.80 & 41.25 & 41.48\\
    \midrule
        AVE     & 32.41 & 41.48 & 42.23 & 42.19 & \textbf{42.27} & 42.11 & 41.56 & 41.61 & 41.47 & 41.39\\
    \bottomrule
    \end{tabular}
    \caption{F1 scores for Span-ASTE + AT. All the results are obtained on the development set by an average of 5 random runs. We highlight the best average results in bold.}
    \label{tab:span-at}
\end{table*}

%% file: tables/bmrc_at.tex
\begin{table*}[]
    \small
    \centering
    \begin{tabular}{ccccccccccc}
    \toprule
         $\alpha$ & 1 & 10 & 30 & 50 & 100 & 500 & 700 & 800 & 1000 & 1500  \\
    \midrule
         E$\to$K & 0.75  & 2.22  & 6.18  & 16.90 & 6.61  & 28.08 & 31.86 & 31.67 & 30.64 & 30.23\\
         B$\to$K & 0.30  & 1.04  & 9.57  & 17.53 & 26.56 & 27.93 & 28.62 & 27.62 & 29.46 & 29.88\\
         F$\to$K & 0.22  & 7.66  & 8.05  & 14.82 & 27.14 & 30.34 & 32.09 & 30.72 & 31.21 & 31.09\\
         H$\to$K & 0.37  & 9.81  & 11.22 & 18.96 & 22.74 & 26.66 & 27.92 & 31.23 & 28.52 & 28.93\\
    \midrule
         E$\to$G & 2.43  & 12.42 & 24.97 & 19.98 & 35.43 & 38.71 & 39.14 & 39.48 & 39.05 & 38.13\\
         B$\to$G & 2.13  & 20.84 & 33.99 & 32.90 & 32.63 & 33.92 & 33.84 & 32.88 & 32.89 & 32.78\\
         F$\to$G & 2.42  & 37.73 & 27.40 & 34.71 & 38.03 & 38.98 & 37.99 & 37.56 & 38.62 & 38.20\\
         H$\to$G & 4.55  & 30.18 & 32.95 & 37.50 & 37.72 & 37.62 & 37.86 &  38.48 & 37.65 & 37.85\\
    \midrule
         E$\to$T & 4.57  & 31.09 & 35.18 & 38.12 & 42.73 & 43.35 & 42.74 & 42.86 & 43.01 & 42.59\\
         B$\to$T & 1.65  & 7.45  & 23.93 & 33.70 & 35.53 & 36.41 & 35.39 & 35.46 & 35.90 & 35.73\\
         F$\to$T & 4.10  & 37.50 & 38.43 & 40.38 & 41.23 & 42.66 & 41.65 & 42.26 & 42.21 & 42.58\\
         H$\to$T & 5.11  & 35.05 & 41.77 & 43.45 & 43.22 & 44.54 & 44.43 & 44.27 & 43.79 & 44.72\\
    \midrule
         E$\to$P & 10.47 & 29.64 & 37.53 & 36.99 & 38.14 & 37.62 & 37.78 & 38.59 & 38.03 & 37.65\\
         B$\to$P & 3.15  & 26.35 & 30.14 & 32.04 & 31.64 & 31.55 & 32.29 & 32.44 & 33.25 & 32.20\\
         F$\to$P & 8.24  & 33.79 & 36.40 & 37.27 & 36.35 & 36.72 & 37.20 & 36.79 & 36.82 & 37.19\\
         H$\to$P & 17.51 & 32.01 & 39.06 & 37.24 & 37.80 & 38.30 & 38.33 & 38.87 & 39.24 & 39.29\\
    \midrule
         AVE     & 4.25  & 22.17 & 27.30 & 30.78 & 33.34 & 35.84 & 36.20 & \textbf{36.32} & 36.27 & 36.19\\
    \bottomrule
    \end{tabular}
    \caption{F1 scores for BMRC + AT. All the results are obtained on the development set by an average of 5 random runs. We highlight the best average results in bold.}
    \label{tab:bmrc-at}
\end{table*}

%% file: acl_latex.bbl
\begin{thebibliography}{30}
\expandafter\ifx\csname natexlab\endcsname\relax\def\natexlab#1{#1}\fi

\bibitem[{Barnes et~al.(2018)Barnes, Badia, and
  Lambert}]{barnes-etal-2018-multibooked}
Jeremy Barnes, Toni Badia, and Patrik Lambert. 2018.
\newblock \href {https://aclanthology.org/L18-1104} {{M}ulti{B}ooked: A corpus
  of {B}asque and {C}atalan hotel reviews annotated for aspect-level sentiment
  classification}.
\newblock In \emph{Proceedings of the Eleventh International Conference on
  Language Resources and Evaluation ({LREC} 2018)}, Miyazaki, Japan. European
  Language Resources Association (ELRA).

\bibitem[{Benesty et~al.(2009)Benesty, Chen, Huang, and
  Cohen}]{benesty2009pearson}
Jacob Benesty, Jingdong Chen, Yiteng Huang, and Israel Cohen. 2009.
\newblock Pearson correlation coefficient.
\newblock In \emph{Noise reduction in speech processing}, pages 1--4. Springer.

\bibitem[{Cai et~al.(2021)Cai, Xia, and Yu}]{cai-etal-2021-aspect}
Hongjie Cai, Rui Xia, and Jianfei Yu. 2021.
\newblock \href {https://doi.org/10.18653/v1/2021.acl-long.29}
  {Aspect-category-opinion-sentiment quadruple extraction with implicit aspects
  and opinions}.
\newblock In \emph{Proceedings of the 59th Annual Meeting of the Association
  for Computational Linguistics and the 11th International Joint Conference on
  Natural Language Processing (Volume 1: Long Papers)}, pages 340--350, Online.
  Association for Computational Linguistics.

\bibitem[{Chen et~al.(2021)Chen, Wang, Liu, and Wang}]{chen2021bidirectional}
Shaowei Chen, Yu~Wang, Jie Liu, and Yuelin Wang. 2021.
\newblock Bidirectional machine reading comprehension for aspect sentiment
  triplet extraction.
\newblock In \emph{Proceedings Of The AAAI Conference On Artificial
  Intelligence}, volume~35, pages 12666--12674.

\bibitem[{Esuli et~al.(2008)Esuli, Sebastiani, and
  Urciuoli}]{esuli-etal-2008-annotating}
Andrea Esuli, Fabrizio Sebastiani, and Ilaria Urciuoli. 2008.
\newblock \href
  {http://www.lrec-conf.org/proceedings/lrec2008/pdf/566_paper.pdf} {Annotating
  expressions of opinion and emotion in the {I}talian content annotation bank}.
\newblock In \emph{Proceedings of the Sixth International Conference on
  Language Resources and Evaluation ({LREC}'08)}, Marrakech, Morocco. European
  Language Resources Association (ELRA).

\bibitem[{Fan et~al.(2019)Fan, Wu, Dai, Huang, and Chen}]{fan2019target}
Zhifang Fan, Zhen Wu, Xinyu Dai, Shujian Huang, and Jiajun Chen. 2019.
\newblock Target-oriented opinion words extraction with target-fused neural
  sequence labeling.
\newblock In \emph{Proceedings of the 2019 Conference of the North American
  Chapter of the Association for Computational Linguistics: Human Language
  Technologies, Volume 1 (Long and Short Papers)}, pages 2509--2518.

\bibitem[{Fei et~al.(2021)Fei, Ren, Zhang, and Ji}]{fei2021nonautoregressive}
Hao Fei, Yafeng Ren, Yue Zhang, and Donghong Ji. 2021.
\newblock Nonautoregressive encoder-decoder neural framework for end-to-end
  aspect-based sentiment triplet extraction.
\newblock \emph{IEEE Transactions on Neural Networks and Learning Systems}.

\bibitem[{Ganin et~al.(2016)Ganin, Ustinova, Ajakan, Germain, Larochelle,
  Laviolette, Marchand, and Lempitsky}]{ganin2016domain}
Yaroslav Ganin, Evgeniya Ustinova, Hana Ajakan, Pascal Germain, Hugo
  Larochelle, Fran{\c{c}}ois Laviolette, Mario Marchand, and Victor Lempitsky.
  2016.
\newblock Domain-adversarial training of neural networks.
\newblock \emph{The journal of machine learning research}, 17(1):2096--2030.

\bibitem[{Guo et~al.(2018)Guo, Shah, and Barzilay}]{guo-etal-2018-multi}
Jiang Guo, Darsh Shah, and Regina Barzilay. 2018.
\newblock \href {https://doi.org/10.18653/v1/D18-1498} {Multi-source domain
  adaptation with mixture of experts}.
\newblock In \emph{Proceedings of the 2018 Conference on Empirical Methods in
  Natural Language Processing}, pages 4694--4703, Brussels, Belgium.
  Association for Computational Linguistics.

\bibitem[{Hussein(2018)}]{hussein2018survey}
Doaa Mohey El-Din~Mohamed Hussein. 2018.
\newblock A survey on sentiment analysis challenges.
\newblock \emph{Journal of King Saud University-Engineering Sciences},
  30(4):330--338.

\bibitem[{Li et~al.(2019)Li, Bing, Li, and Lam}]{li2019unified}
Xin Li, Lidong Bing, Piji Li, and Wai Lam. 2019.
\newblock A unified model for opinion target extraction and target sentiment
  prediction.
\newblock In \emph{Proceedings of the AAAI conference on artificial
  intelligence}, volume~33, pages 6714--6721.

\bibitem[{Liu et~al.(2021)Liu, Xu, Yu, Dai, Ji, Cahyawijaya, Madotto, and
  Fung}]{liu2021crossner}
Zihan Liu, Yan Xu, Tiezheng Yu, Wenliang Dai, Ziwei Ji, Samuel Cahyawijaya,
  Andrea Madotto, and Pascale Fung. 2021.
\newblock Crossner: Evaluating cross-domain named entity recognition.
\newblock In \emph{Proceedings of the AAAI Conference on Artificial
  Intelligence}, volume~35, pages 13452--13460.

\bibitem[{Mao et~al.(2021)Mao, Shen, Yu, and Cai}]{mao2021joint}
Yue Mao, Yi~Shen, Chao Yu, and Longjun Cai. 2021.
\newblock A joint training dual-mrc framework for aspect based sentiment
  analysis.
\newblock In \emph{Proceedings of the AAAI Conference on Artificial
  Intelligence}, volume~35, pages 13543--13551.

\bibitem[{Mukherjee et~al.(2021)Mukherjee, Nayak, Butala, Bhattacharya, and
  Goyal}]{mukherjee-etal-2021-paste}
Rajdeep Mukherjee, Tapas Nayak, Yash Butala, Sourangshu Bhattacharya, and Pawan
  Goyal. 2021.
\newblock \href {https://doi.org/10.18653/v1/2021.emnlp-main.731} {{PASTE}: A
  tagging-free decoding framework using pointer networks for aspect sentiment
  triplet extraction}.
\newblock In \emph{Proceedings of the 2021 Conference on Empirical Methods in
  Natural Language Processing}, pages 9279--9291, Online and Punta Cana,
  Dominican Republic. Association for Computational Linguistics.

\bibitem[{Ni et~al.(2019)Ni, Li, and McAuley}]{ni-etal-2019-justifying}
Jianmo Ni, Jiacheng Li, and Julian McAuley. 2019.
\newblock \href {https://doi.org/10.18653/v1/D19-1018} {Justifying
  recommendations using distantly-labeled reviews and fine-grained aspects}.
\newblock In \emph{Proceedings of the 2019 Conference on Empirical Methods in
  Natural Language Processing and the 9th International Joint Conference on
  Natural Language Processing (EMNLP-IJCNLP)}, pages 188--197, Hong Kong,
  China. Association for Computational Linguistics.

\bibitem[{Peng et~al.(2020)Peng, Xu, Bing, Huang, Lu, and Si}]{peng2020knowing}
Haiyun Peng, Lu~Xu, Lidong Bing, Fei Huang, Wei Lu, and Luo Si. 2020.
\newblock Knowing what, how and why: A near complete solution for aspect-based
  sentiment analysis.
\newblock In \emph{Proceedings of the AAAI Conference on Artificial
  Intelligence}, volume~34, pages 8600--8607.

\bibitem[{Pontiki et~al.(2016)Pontiki, Galanis, Papageorgiou, Androutsopoulos,
  Manandhar, Al-Smadi, Al-Ayyoub, Zhao, Qin, De~Clercq
  et~al.}]{pontiki2016semeval}
Maria Pontiki, Dimitrios Galanis, Haris Papageorgiou, Ion Androutsopoulos,
  Suresh Manandhar, Mohammad Al-Smadi, Mahmoud Al-Ayyoub, Yanyan Zhao, Bing
  Qin, Orph{\'e}e De~Clercq, et~al. 2016.
\newblock Semeval-2016 task 5: Aspect based sentiment analysis.
\newblock In \emph{International workshop on semantic evaluation}, pages
  19--30.

\bibitem[{Pontiki et~al.(2015)Pontiki, Galanis, Papageorgiou, Manandhar, and
  Androutsopoulos}]{pontiki2015semeval}
Maria Pontiki, Dimitrios Galanis, Harris Papageorgiou, Suresh Manandhar, and
  Ion Androutsopoulos. 2015.
\newblock Semeval-2015 task 12: Aspect based sentiment analysis.
\newblock In \emph{Proceedings of the 9th international workshop on semantic
  evaluation (SemEval 2015)}, pages 486--495.

\bibitem[{Pontiki et~al.(2014)Pontiki, Galanis, Pavlopoulos, Papageorgiou,
  Androutsopoulos, and Manandhar}]{pontiki-etal-2014-semeval}
Maria Pontiki, Dimitris Galanis, John Pavlopoulos, Harris Papageorgiou, Ion
  Androutsopoulos, and Suresh Manandhar. 2014.
\newblock \href {https://doi.org/10.3115/v1/S14-2004} {{S}em{E}val-2014 task 4:
  Aspect based sentiment analysis}.
\newblock In \emph{Proceedings of the 8th International Workshop on Semantic
  Evaluation ({S}em{E}val 2014)}, pages 27--35, Dublin, Ireland. Association
  for Computational Linguistics.

\bibitem[{Poria et~al.(2014)Poria, Cambria, Ku, Gui, and
  Gelbukh}]{poria2014rule}
Soujanya Poria, Erik Cambria, Lun-Wei Ku, Chen Gui, and Alexander Gelbukh.
  2014.
\newblock A rule-based approach to aspect extraction from product reviews.
\newblock In \emph{Proceedings of the second workshop on natural language
  processing for social media (SocialNLP)}, pages 28--37.

\bibitem[{Rosenstein et~al.(2005)Rosenstein, Marx, Kaelbling, and
  Dietterich}]{Rosenstein2005}
Michael~T Rosenstein, Zvika Marx, Leslie~Pack Kaelbling, and Thomas~G
  Dietterich. 2005.
\newblock To transfer or not to transfer.
\newblock \emph{NIPS}.

\bibitem[{Tang and Jia(2020)}]{tang2020discriminative}
Hui Tang and Kui Jia. 2020.
\newblock Discriminative adversarial domain adaptation.
\newblock In \emph{Proceedings of the AAAI Conference on Artificial
  Intelligence}, volume~34, pages 5940--5947.

\bibitem[{Wu et~al.(2020)Wu, Ying, Zhao, Fan, Dai, and Xia}]{wu-etal-2020-grid}
Zhen Wu, Chengcan Ying, Fei Zhao, Zhifang Fan, Xinyu Dai, and Rui Xia. 2020.
\newblock \href {https://doi.org/10.18653/v1/2020.findings-emnlp.234} {Grid
  tagging scheme for aspect-oriented fine-grained opinion extraction}.
\newblock In \emph{Findings of the Association for Computational Linguistics:
  EMNLP 2020}, pages 2576--2585, Online. Association for Computational
  Linguistics.

\bibitem[{Xu et~al.(2021)Xu, Chia, and Bing}]{xu-etal-2021-learning}
Lu~Xu, Yew~Ken Chia, and Lidong Bing. 2021.
\newblock \href {https://doi.org/10.18653/v1/2021.acl-long.367} {Learning
  span-level interactions for aspect sentiment triplet extraction}.
\newblock In \emph{Proceedings of the 59th Annual Meeting of the Association
  for Computational Linguistics and the 11th International Joint Conference on
  Natural Language Processing (Volume 1: Long Papers)}, pages 4755--4766,
  Online. Association for Computational Linguistics.

\bibitem[{Xu et~al.(2020)Xu, Li, Lu, and Bing}]{xu-etal-2020-position}
Lu~Xu, Hao Li, Wei Lu, and Lidong Bing. 2020.
\newblock \href {https://doi.org/10.18653/v1/2020.emnlp-main.183}
  {Position-aware tagging for aspect sentiment triplet extraction}.
\newblock In \emph{Proceedings of the 2020 Conference on Empirical Methods in
  Natural Language Processing (EMNLP)}, pages 2339--2349, Online. Association
  for Computational Linguistics.

\bibitem[{Yan et~al.(2021)Yan, Dai, Ji, Qiu, and Zhang}]{yan-etal-2021-unified}
Hang Yan, Junqi Dai, Tuo Ji, Xipeng Qiu, and Zheng Zhang. 2021.
\newblock \href {https://doi.org/10.18653/v1/2021.acl-long.188} {A unified
  generative framework for aspect-based sentiment analysis}.
\newblock In \emph{Proceedings of the 59th Annual Meeting of the Association
  for Computational Linguistics and the 11th International Joint Conference on
  Natural Language Processing (Volume 1: Long Papers)}, pages 2416--2429,
  Online. Association for Computational Linguistics.

\bibitem[{Zhang et~al.(2020)Zhang, Li, Song, and
  Wang}]{zhang-etal-2020-multi-task}
Chen Zhang, Qiuchi Li, Dawei Song, and Benyou Wang. 2020.
\newblock \href {https://doi.org/10.18653/v1/2020.findings-emnlp.72} {A
  multi-task learning framework for opinion triplet extraction}.
\newblock In \emph{Findings of the Association for Computational Linguistics:
  EMNLP 2020}, pages 819--828, Online. Association for Computational
  Linguistics.

\bibitem[{Zhang et~al.(2021{\natexlab{a}})Zhang, Deng, Li, Yuan, Bing, and
  Lam}]{zhang-etal-2021-aspect-sentiment}
Wenxuan Zhang, Yang Deng, Xin Li, Yifei Yuan, Lidong Bing, and Wai Lam.
  2021{\natexlab{a}}.
\newblock \href {https://doi.org/10.18653/v1/2021.emnlp-main.726} {Aspect
  sentiment quad prediction as paraphrase generation}.
\newblock In \emph{Proceedings of the 2021 Conference on Empirical Methods in
  Natural Language Processing}, pages 9209--9219, Online and Punta Cana,
  Dominican Republic. Association for Computational Linguistics.

\bibitem[{Zhang et~al.(2021{\natexlab{b}})Zhang, Li, Deng, Bing, and
  Lam}]{zhang2021towards}
Wenxuan Zhang, Xin Li, Yang Deng, Lidong Bing, and Wai Lam. 2021{\natexlab{b}}.
\newblock Towards generative aspect-based sentiment analysis.
\newblock In \emph{Proceedings of the 59th Annual Meeting of the Association
  for Computational Linguistics and the 11th International Joint Conference on
  Natural Language Processing (Volume 2: Short Papers)}, pages 504--510.

\bibitem[{Zhang et~al.(2021{\natexlab{c}})Zhang, Guo, and
  Kordjamshidi}]{zhang-etal-2021-towards}
Yue Zhang, Quan Guo, and Parisa Kordjamshidi. 2021{\natexlab{c}}.
\newblock \href {https://doi.org/10.18653/v1/2021.splurobonlp-1.5} {Towards
  navigation by reasoning over spatial configurations}.
\newblock In \emph{Proceedings of Second International Combined Workshop on
  Spatial Language Understanding and Grounded Communication for Robotics},
  pages 42--52, Online. Association for Computational Linguistics.

\end{thebibliography}
